\documentclass{article}
\usepackage{preamble}

\title{Machine-Assisted Grading of Nationwide School-Leaving Essay Exams with LLMs and Statistical NLP}
\author{
Andres Karjus\textsuperscript{1,2,3}\hspace{0.2mm} 
Kais Allkivi\textsuperscript{1}\hspace{0.2mm} 
Silvia Maine\textsuperscript{1}\hspace{0.2mm} 
Katarin Leppik\textsuperscript{1}\hspace{0.2mm} 
Krister Kruusmaa\textsuperscript{1,4}\hspace{0.2mm} 
Merilin Aruvee\textsuperscript{1} \\
\textsuperscript{1}Tallinn University \hspace{0.7mm} \textsuperscript{2}University of Tartu \hspace{0.7mm}  \textsuperscript{3}Estonian Business School \hspace{0.7mm}  \textsuperscript{4}Institute of the Estonian Language
}
\date{\vspace{-1em}}

\begin{document}
\maketitle
\begin{abstract}
Large language models (LLMs) enable rapid and consistent automated evaluation of open-ended exam responses, including dimensions of content and argumentation that have traditionally required human judgment. This is particularly important in cases where a large amount of exams need to be graded in a limited time frame, such as nation-wide graduation exams in various countries. Here, we examine the applicability of automated scoring on two large datasets of trial exam essays of two full national cohorts from Estonia. We operationalize the official curriculum-based rubric and compare LLM and statistical natural language processing (NLP) based assessments with human panel scores. The results show that automated scoring can achieve performance comparable to that of human raters and tends to fall within the human scoring range. We also evaluate bias, prompt injection risks, and LLMs as essay writers. These findings demonstrate that a principled, rubric-driven, human-in-the-loop scoring pipeline is viable for high-stakes writing assessment, particularly relevant for digitally advanced societies like Estonia, which is about to adapt a fully electronic examination system. Furthermore, the system produces fine-grained subscore profiles that can be used to generate systematic, personalized feedback for instruction and exam preparation. The study provides evidence that LLM-assisted assessment can be implemented at a national scale, even in a small-language context, while maintaining human oversight and compliance with emerging educational and regulatory standards.

\end{abstract}

\section{Introduction}

The rapid development of artificial intelligence (AI), and especially large language models (LLMs), has renewed interest in how open-ended student writing can be evaluated at scale. Essay scoring is one of the most demanding components of educational assessment: human raters must interpret complex rubrics, ensure consistent judgments, and process large volumes of text under time constraints. These challenges are amplified in small-language contexts, where commercial automated scoring tools are scarce and human evaluator pools limited. Estonia is one such small country, yet widely recognized for its advanced digital public infrastructure and ranking high in international digital government indices \parencite{solvak_case_2024,kattel_estonias_2019}, with digital availability of government services having reached 100\% coverage in 2025 \parencite{kriisa_estonia_2025}. The country is also in the process of modernizing its basic and secondary education examination infrastructure, with the goal to move to nationwide computer-based graduation exams by 2027 \parencite{education_estonia_estonias_2023,eurydice_national_2025}. The move to e-exams is expected to support more objective scoring, increased efficiency in administration, and the use of learning analytics \parencite{eurydice_national_2025}. Born-digital, machine-readable texts enable easier automatic assessment, but this requires systematic evaluation, transparent practices, and compliance with European laws and regulations. At the same time, the process of integrating AI tools into the school system and curriculum as a matter of national policy is ongoing, dubbed the "AI Leap" \parencite{ai_leap_ai_2025,ministry_of_education_and_research_estonia_2025}.

Automated Essay Scoring (AES) has been studied since the 1960s \parencite{page1968use}, and subsequent work has emphasized two central psychometric requirements: reliability (consistency of scoring) and validity (alignment with intended constructs and educational goals) \parencite{attali2013}. Early AES systems offered mechanical consistency but captured mainly surface-level features, falling short on deeper aspects of argumentation, evidence use, or reasoning. Efforts to improve these systems have combined embeddings, neural architectures and feature-engineered linguistic indicators \parencite{zehner_automatically_2018,catulay_neural-network_2021,ramesh_automated_2022,elmassry_systematic_2025}. Yet, transparency, trust, and interpretability remain persistent concerns, especially in high-stakes contexts like examinations that determine future opportunities of the test-takers \parencite{conijn_effects_2023}.

State-of-the-art LLMs can approximate human scoring on complex writing tasks, particularly when evaluation is closely aligned with rubric descriptors \parencite{wang_evaluating_2025,elmassry_systematic_2025,jung_combining_2024}, as well as generate examination materials \parencite{ripoll_y_schmitz_evaluating_2025}.
Reliability varies across tasks and configurations: some models show strong inter-rater stability and acceptable validity, but performance can fluctuate across updates \parencite{packLargeLanguageModels2024}, models may exhibit conservative scoring tendencies or a positivity bias observed also in other LLM applications \parencite{sessler2024CanAIGrade,kundu2024AreLargeLanguage}, and evaluations of disciplinary writing reveal tendencies to reward surface fluency over deeper reasoning \parencite{olivos2024instructor,lundgren2024LargeLanguageModels}. Similar patterns appear in short-answer scoring, where LLMs perform well on clear-cut cases but require human oversight for borderline responses \parencite{xiao2024}. 
The variability across models, prompts, and domains underscores the need for controlled, rubric-anchored workflows rather than unconstrained zero-shot prompting.

LLM behavior is also sensitive to prompt phrasing, framing, and adversarial variation: small changes in structure or task description can shift scores substantially, raising concerns about robustness in high-stakes contexts \parencite{yamamura2024fooling,zhou2024,mavridis2023}. Beyond scoring, LLMs can support formative feedback, though feedback may become formulaic or lose pedagogical nuance without strong rubric grounding \parencite{nguyen2024,deng2023}. Long-term perspectives caution that while automated scoring scales well, it can erode teacher control if not embedded in clearly defined human-in-the-loop processes \parencite{dominguez2023}. At the same time, LLMs show promise as writing scaffolds and coaching tools that improve revision quality and learning outcomes when used appropriately \parencite{ramesh_automated_2022,kostic2024llms}.

Concerns about fairness and algorithmic bias remain central. Frameworks for assessing and reducing bias in educational evaluation emphasize the need for systematic auditing \parencite{wang2024adversarial}. Acceptance and legitimacy also depend on perceived fairness: student willingness to trust AI-generated grades is tied to transparency and consistency \parencite{najafi_ai-assisted_2025,jones-jang_fairness_2025}. Evidence from operational large-scale tests, such as the fully automated scoring of PTE Academic, illustrates this duality: test-takers value consistency yet question whether automated systems capture nuanced performance \parencite{leaton_gray_power_2025}. Similar observations appear in speech-recognition-based oral exam scoring, where internally consistent systems nonetheless showed leniency biases \parencite{xu_assessing_2021}, reflecting challenges analogous to those seen in current pretrained LLMs \parencite{rrv_chaos_2024}. However, these are not insurmountable challenges as machine learning models have been routinely investigated for bias for a long time now in computational social science and artificial cognition fields, and various methods exist to do so systematically \parencite{taylor_artificial_2021,acerbi_large_2023,begus_experimental_2024,kotek_gender_2023}.

\subsection{Regulatory context and implications for national assessment}

Technical advances in AES and LLM-based scoring coincide with an evolving policy and regulatory environment. The EU AI Act adopts a risk-based framework in which AI systems used to evaluate learning outcomes, determine access or admission to education, or assign individuals to educational pathways are classified as high-risk \parencite[Annex III, ][]{eu_regulation_2024}. Such systems are subject to stringent requirements on risk management, data governance, transparency, documentation, and human oversight.

The US Office of Educational Technology’s report "Artificial Intelligence and the Future of Teaching and Learning" emphasizes that AI in education should be designed with "humans in the loop", with educators retaining decision authority and AI systems being inspectable, explainable and overridable, and allowing scoring to remain aligned with pedagogical aims \parencite{cardona_artificial_2023}.
This echoes various recent academic position papers calling for responsible design and oversight \parencite{ozmen_garibay_six_2023}.

For national examination systems in EU member states such as Estonia, these provisions imply that fully automated grading of high-stakes exams, without meaningful human control, is not acceptable. Machine-assisted scoring must operate within a human-in-the-loop architecture in which teachers or trained assessors retain responsibility for final scores, and AI outputs are treated as decision support rather than binding decisions. Estonia’s broader digital governance model and high public trust in these institutions \parencite{solvak_case_2024,oecd_government_2025} provide favorable conditions for implementing such regulated AI systems. The same infrastructure that supports proactive and personalized public services can, in principle, support transparent and auditable exam-scoring pipelines, provided that data protection and educational equity requirements are met.

\subsection{The Estonian national essay as a test case}

The Estonian school system consists of two main components, basic school (years 1-9) and upper secondary school (years 10-12), with the latter branching into general and vocational education. Both the basic and secondary school end with nationally standardized examinations in several compulsory and optional subjects, with the Estonian language being one of the former.

Essay writing has been a central part of Estonian as a First Language (L1) education for decades and became a formal component of the national examination system in 1997 \parencite{lepajoeKirjandKuiTekstiliik2011}. The essay (\textit{kirjand}) format has changed over the years, from a longer essay to a combination of reading comprehension task with a shorter essay, nominally 200 words in basic school and 400 in upper secondary school. Compared to the basic school exam, the 12th grade essays have higher expectations for abstraction, argumentation, and intertextual reference.

Essay topics are typically socially relevant and age-appropriate, such as career choices, the future of the Estonian language, or healthy eating habits among youth. Students are given a topic and are expected to produce an argumentative essay, which is assessed using a multi-trait holistic scoring model, covering content, spelling and grammar, vocabulary and style, and structural organization.
Scores on these exams are consequential for progression and access to further education; upper secondary exams are centrally assessed, while basic school exams are evaluated locally by teachers but under nationally defined criteria that are accompanied by guidelines and a rubric for evaluation. Both exams are carried out by the Education and Youth Board (Haridus- ja Noorteamet, HarNo), a government agency of the Ministry of Education and Research.

The upper secondary exam incorporates a dual-rater blind review system. After the evaluation of two experts, the essay receives a grade, which is the mean of two scores. If the discrepancy of the two scores is more than 30\%, a third review is conducted. In 2023, 5.5\% of 7,220 exam essays underwent a third review, and 10 required a fourth. In 2024, of the 7,672 examinees, 8.5\% essays needed a third assessment, with seven receiving a fourth. This rate of multi-stage grading, along with a notable number of successful appeals: 51\% of 69 appeals in 2023 and 60\% of 96 in 2024 resulted in score increases (data provided to the authors directly by HarNo) indicates a certain degree of inconsistency in grading. These findings suggest that even with standardized rubrics and multiple reviews, the current human-based system does not fully ensure objectivity or reliability.

While the genre is intended to develop coherent expression and argumentation skills, it has been criticized as detached from functional, purpose-driven language use. To date, the exam task has remained unchanged since 2013, but have undergone some minor phrasing adjustments over the years, which haven't altered the content of the rubrics. However, with the transition to a digital exam format, its content will undergo modification in the coming years. The current fully manual workflows limit the feasibility of providing detailed feedback to students beyond a single numeric score. These and related issues have drawn public critique in the local media \parencite{saarniit_harnos_2025,veelmaa_pekske_2025,rudi_koolijuhid_2025} and in academic papers \parencite{kapp2024kirjutan, kapp2024mida}.
This mirrors academic discourse of essay-based exams as both biased for assessing higher-order language skills and vulnerable to subjectivity and inconsistency in scoring \parencite{brown_reliability_2009,hasan_assessing_2024,ramesh_automated_2022}.

Work on Estonian as L1 digital examination system began in 2014 \parencite{jakobson_keel_2024}. The initial research report, already produced in 2021, informed the development of the e-assessment framework \parencite{aruvee_eesti_2021}. In 2022, an e-exam writing task and its accompanying guidelines along with an evaluation rubric were devised. Pilot e-exams have been conducted every spring since 2023, involving teachers and national assessors. Their feedback has led to clearer evaluation criteria and more detailed rubrics. The focus has been primarily on reforming the basic school exam, with adjustments to the upper secondary exam in the future. Current national plans are to implement basic school e-exam in year 2027, and upper secondary exam in forthcoming years. Given systemic changes and long-standing concerns around subjectivity, the transition to digital assessment raises an important question: could artificial intelligence contribute to evaluating these essays more fairly or at least more efficiently? As digital exams in Estonia will be administered through the Examination Information System (EIS), with some task types, such as reading and listening comprehension, already graded automatically, the potential to use LLMs in AES appears viable. Incorporating AI in grading, even partially, could reduce assessor workload and potentially improve consistency.

\subsection{Positioning this study}

This combination of digital delivery, structured rubrics, national relevance, and public openness to digital solutions and AI in education makes the Estonian essay exam an informative test case for exploring LLM-assisted scoring in high-stakes setting, as well for exploring the applicability of multilingual LLMs in smaller, lower-resource languages for such tasks. Low resources here refer to the availability of training corpora of machine-readable texts and other natural language processing (NLP) resources. Multilingual LLMs rely on being trained on a wealth of such data, which might not be available (at least for companies training LLMs) in languages with fewer speakers or a lower digital presence on the web (Estonia is a country of 1.3 million; the Estonian language has about 900,000 native speakers).
 
This study follows the aforementioned research on evaluating applicability and reliability of machine-assisted grading on the example of pretrained LLMs, used in a zero-shot in-context learning setting, where the model input consists of the rubric and the essay to be evaluated. We use the complete national cohort of 1,559 digitally produced argumentative essays from the (already digitally delivered) trial exam session in 2024, acquired directly from HarNo. We then employ state-of-the-art LLMs (OpenAI GPT and Google Gemini models), compare their outputs with the official human scores, and sketch a framework for systematic integration of machine-assisted grading for exams.

In doing so, the study contributes to ongoing debates on the role of LLMs in educational assessment by providing evidence from a high-stakes national-scale exam. The results illustrate how LLMs and language technology can be integrated into a human-in-the-loop scoring workflow that supports faster and more consistent evaluation, while also generating fine-grained subscore profiles that can be used for systematic, rubric-aligned feedback in instruction and exam preparation. Within Estonia’s broader digital and AI-in-education strategy, the case demonstrates a concrete pathway for embedding LLM-assisted assessment into national examination practice without relinquishing human judgment or regulatory compliance.

Some caveats. Importantly, this is not an LLM benchmarking study: our goal is not to determine which current LLM is best at scoring essays written in Estonian, but rather to provide a proof of concept, showing that such a task can be feasibly automated using technology readily available already today. It is also crucial to distinguish similar yet distinct uses of the same technology: LLMs like any other machine learning models can be used directly (locally or via cloud services), but also form the central components of modern chatbots like ChatGPT, Gemini, or Copilot. The discussion below pertains only to the usage of LLMs (as machine grading tools), and not chatbots, which are for many reasons often not the optimal solution for large scale data analysis as well as grading tasks \parencite[see][]{karjus_machine-assisted_2025}.

\section{Methods and materials}

\subsection{Data: Estonian trial exam essays}

HarNo, the Estonian Education and Youth Board (\textit{Haridus- ja Noorteamet}), provided e-exam essays from the trial run of spring 2024 for this study, 781 texts from 9th year basic school students and 764 texts from 12th year upper secondary students. Each text is assessed by two assessors (the full year 9 dataset contained 795 essays, but 14 had only one grader, so these were excluded). 

\begin{figure}[hbt]
	\noindent
	\includegraphics[width=\columnwidth]{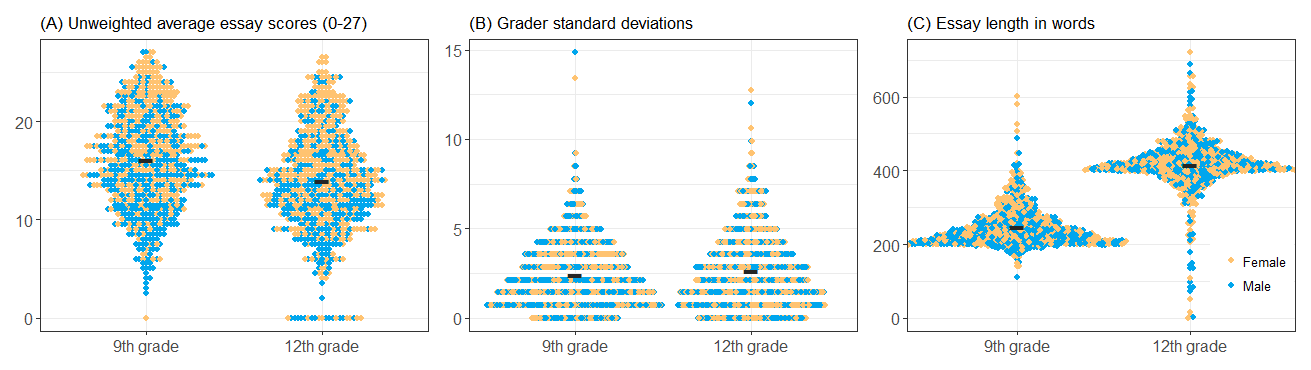}
	\caption{
Overview of the two datasets: average score, standard deviations between grader scores, and essay lengths, as "beeswarm" graphs where points are pushed aside to avoid overlap, illustrating the distributions. Each essay is one dot, colored by student sex (as this was available in the metadata). Black bars are averages. With the exception of some outliers, the samples are fairly realistic in terms of scores, what is known about grading challenges in terms of inter-grader agreement, and length, with most students following the guidelines of their respective tasks (therefore, longer texts in 12th grade).
    }\label{fig_summary}
\end{figure}

As visible in Figure \ref{fig_summary}, some essays received a zero score - these are mostly cases where the student did not take the (trial) exam seriously and wrote only a few words or just the title, composed a poem, or something irrelevant for the task. We did not remove any such outliers and analyzed the entire dataset. The largest discrepancy between assessors is a case where one assigned 19 points out of 27 and the second a zero, for a simplistic albeit grammatically mostly correct text (which may also be a dataset error; see the yellow point in the top left of Figure \ref{fig_summary}.B).

\subsection{Evaluation rubrics }

The most recent evaluation rubric for 9th grade approved by HarNo since 2023 consists of three main aspects: (1) content and structure (12 points), assessed through four criteria: title and introduction, use of source material, argumentation, and conclusion; (2) vocabulary and syntax (6 points), split into two respective criteria focusing on lexical choices and sentence structure (e.g., word order, agreement, government); and (3) orthography and formatting (9 points) divided into three criteria: orthography and morphology (word formation), punctuation, and structuring and formatting covering overall structural consistency and typographical correctness. The rubric essentially operationalizes the L1 national curriculum, translating the learning outcomes of critical thinking, textual interpretation, and effective communication into assessable criteria.

Each of the nine categories is graded on a 0 to 3 scale (but weighted later), where 3 implies perfect or near-perfect execution, and the score is progressively decremented depending on the observed mistakes. Maximal score expects a text with a generalizable argument relevant to the topic, clearly and convincingly presented. Lower scores reflect incoherence and lack of clarity. Spelling is assessed based on the number of errors, while style and structure are evaluated based on criteria referring to the essay as a whole (see the Appendix for the full grading rubric).

Both the 9th and 12th grade essay evaluation rubrics follow the same 27-point structure internally, but the scores are weighted before calculating the final score visible to the student, giving more weight to content categories. The main difference between the two exams lies in expectations: the 9th grade model emphasizes shorter texts with at least one integrated source and basic argumentative coherence, while the 12th grade model requires longer essays, integration of at least two sources, more sustained reasoning, richer language use, and stricter accuracy. This progression reflects a shift from foundational writing skills to advanced academic literacy. 

Both exams are structured similarly with reading and writing sections. In the 12th grade exam, students must choose one of four given prompts and independently formulate a title. The required length is 400 words and the grading model is similar but scaled for increased complexity. The maximum score is 60 points, including 25 for content, 20 for spelling and grammar, 10 for style, and 5 for structure. The content component rewards deep analysis, strong argumentation, generalization and well-chosen examples. The final score for each component is calculated by multiplying the level achieved by a component-specific weight. Specifically, the content score is calculated by multiplying the level (1-5) by 5, spelling and grammar by 4, style by 2, and structure by 1. For example, a level 4 in content results in 20 points (4*5), while the same level in spelling and grammar yields 16 points (4*4). Here we omit the weights for simplicity and report results on the original 0-3 scales (0-27 total; see Appendix for both detailed grading models).

The grading instructions are quite challenging for both humans and machines, requiring unavoidably subjective judgment in some cases like assessing whether the title is "captivating", arguments are "clear" and the vocabulary "rich and unique" (but also not too nonstandard for written language, as that would lower the score). Other categories rely on counting mistakes, but require differentiating between accidental typos and morphological errors, as well as not counting the same "type" of mistake more than once, which requires determining whether two mistakes are in the same category and of the same type or not.

\subsection{Zero-shot learning with LLMs} 

Here we use a very simple zero-shot approach to text analytics that has proven very successful in various research domains engaged in analyzing and classifying texts like humanities and social science fields \parencite{ziems_can_2023,gilardi_chatgpt_2023,qin_is_2023,karjus_machine-assisted_2025}. In short, a generative LLM is presented with a single input consisting of two parts, the instructions --- in this case the evaluation rubric and context of the task --- and the essay to be graded. All models received the same input, instructing them to output a short summary reasoning for the score given (for debugging and qualitative insights), and the score itself.
Here we evaluated each rubric category separately, that is, each essay was submitted to the model nine times with a prompt to only assess the given category. A time and cost optimization strategy would be to submit the entire rubric and the essay once, now made more feasible with the newest frontier models being trained to follow "structured outputs" that constrain the generated text and enable better consistency and easier parsing. However, doing each aspect separately ensures they do not affect each other, and the shorter context window prevents "context rot" in the model \parencite{liu_lost_2024}.

The full prompts are provided in the Appendix. As an example, these are the components of the 9th grade evaluation prompt:
\begin{quote}
\itshape
\setlength{\parindent}{0pt}
Your task is to grade this Estonian 9th grade exam essay kirjand in """triple quotes""" below, using the provided grading rubric. For reference, the student was asked to do the following: [the writing prompt provided to the student]\\
Here you will ONLY grade this aspect: [rubric category] Grade this on a scale of 0 to 3 points. Keep in mind this is just a text by an Estonian 9th grader. Use the following criteria to assess how many points to award, following this principle: if all the criteria are fulfilled then give 3 points, but if something is lacking, then lower points as described in this rubric:

Two alustekst reference texts were provided here; the student is expected to engage with at least one of them. [summaries of the two texts] Take this into account and grade:\\
3 if: ... [the definitions of the grades from 3 to 0]

First explain your reasoning of relevant rubric criteria {[longer format definition for reasoning and grade output]} \\
Essay text begins below:
\end{quote}

Essentially, the LLMs, being instruction-tuned models, are instructed as one would a human grader: they are given context, a task, clear definitions, and the text to analyze.
We evaluated the following models known to be capable in the Estonian language: GPT-4o-2024-05-13, GPT-4.1-2025-04-14, o4-mini-2025-04-16 (with the reasoning parameter set to "low"), Gemini-2.0-Flash, and Gemini-1.5-Pro. All of them belong to the "commercial" or "closed-weights" class of LLMs; while "open-weights" models can in some cases provide better transparency and enable closer scrutiny (due to access to model weights, and potentially, although rarely, also training data), the selection of capable open LLMs for Estonian at the time of conducting the experiments was rather limited. Running open models locally also requires high-performance computing hardware, whereas the aforementioned models can all be conveniently used as cloud services via their respective APIs (application programming interfaces). 

Costs of such services have come down as well. For example, the entire 9th grade dataset consists of 400,433 LLM tokens (the parts of words LLMs operate with) if using the GPT tokenizer; we spent 9,098,937 tokens total due to adding prompts and running each essay 9 times. On GPT-4.1, with the added small cost of outputs, this cost around \$20, while on GPT-5.1, newest in this line of LLMs at the time of writing, it would cost only around a dozen. With the "batch processing" option of this service, the costs would be further halved.
Of course, in real production cases where real exams are at stake, a locally monitored architecture and controlled model would likely be preferable. Here, the aforementioned frontier models serve as useful examples.

As with all current LLM research, model choices and the reported results naturally become outdated the moment they are completed. However, benchmarking is not the goal here, and neither is finding the "best" model for essay scoring. The goal is to explore one class of models as possible machine-assisted grading tools. As shown in results below, the results are rather jagged as expected \parencite{dellacqua_navigating_2023}, but overall quite similar across rubric categories, with no clear "winner". Naturally, other types of tools exist for automatically assessing some if not all the rubric categories, such as various purpose-built machine learning classifier approaches, lexical diversity scores, and spellcheckers as a way to detect mistakes --- but all these tools require either training or some additional statistical prediction to turn their outputs into rubric scores --- whereas LLMs can simply be instructed. Instructions in turn can be iteratively improved as well, which can improve performance; here we did not pursue this direction and report initial exploratory results.

\subsection{Supervised learning using statistical NLP tools}

For comparison with LLMs, we also used feature-based supervised learning to grade the five language structure and correctness related categories of the total nine in the rubric. Unlike using LLMs, this approach requires labeled
training data with expert ratings to predict the scores of new texts. The predictions are based on the relationships between various linguistic features and average subscores of individual rubric aspects. A similar approach has been used to assess the proficiency level of Estonian as a second language (L2) learner writings on the scale of A2–C1 \parencite{vajjala-loo-2014-automatic, allkivi-2025-cefr}. These studies have made use of grammatical, lexical, surface complexity, and error features to train text classification models. However, the task of scoring Estonian L1 essays should rather be handled as a regression problem.

Using NLP tools, we calculated quantitative linguistic features of the essays related to five aspects: \emph{vocabulary} and \emph{syntax}, as well as \emph{punctuation}, \emph{orthography and morphology}, and \emph{structuring and formatting}. 
We extracted 108 features, which can be divided into four categories: grammatical features (53), lexical features (20), surface features that do not require deeper linguistic analysis (12), and error features obtained with spelling and grammar correction tools (23). For the scoring experiments, we associated each feature with one or several of the graded aspects. The linguistic features and feature extraction are described in Appendix A3.\\
\textbf{Error features} relate to all considered grading aspects. We assumed the following associations between the aspects and automated correction types:

\begin{itemize}[topsep=0pt,parsep=0pt]
  \item \emph{Punctuation} – missing, unnecessary, and replaced punctuation corrections
  \item \emph{Orthography and morphology} – spelling, word replacement, whitespace, and mixed corrections (i.e., several edit types co-occur)
  \item \emph{Structuring and formatting} – spelling and word replacement corrections
  \item \emph{Vocabulary} – word replacement, missing/unnecessary word, word order, and mixed corrections
  \item \emph{Syntax} – word order, missing/unnecessary word, and mixed corrections
\end{itemize}

Some features relate to multiple aspects. Spelling corrections can indicate both orthographic and typographical errors, whereas word replacements also occur when the student has chosen an incorrect word or word form. Wording difficulties are reflected in corrections of missing and unnecessary words, as well as word order and mixed corrections. Exceeding short expressions, the same features provide information about problems with sentence structure. Although it is not optimal to use similar features for scoring different aspects, it helped us gain an understanding of the relationship between potentially informative features and particular subscores.

\textbf{Surface features} include text complexity metrics based on word and sentence length, associated with \emph{syntax} and \emph{vocabulary}, but also variables of text structure, i.e., measures of paragraph count and length, relating to \emph{structuring and formatting}.

\textbf{Lexical features} represent different dimensions of vocabulary complexity, i.e., lexical diversity, sophistication, and density (see Read \citeyear{read-2000-vocabulary}; Malvern et al. \citeyear{malvernetal-2004-vocabulary}). Lexical diversity expresses the proportion of unique vocabulary, whereas lexical sophistication comprises the rarity and abstractness of words. Lexical density, i.e., the proportion of content words, has been used to analyze both the vocabulary and syntax of Estonian L1 student writings \parencite{kergeetal-2014-vocabulary, kergeetal-2013-syntax}. The use of lexical and grammatical words affects the style and informativeness of the text, as well as the structure of sentences. Similarly, we assumed that rare and abstract words can entail specific grammatical patterns. We therefore treated lexical density and sophistication features as potential predictors of the \emph{syntax} score in addition to the \emph{vocabulary} score.

\textbf{Grammatical features} can be divided into part of speech (POS), nominal, and verb features, which can all be associated with the aspect of \emph{syntax}. We additionally used POS features to score \emph{vocabulary}, as they relate to forming multi-word expressions that function as lexical units, such as noun phrases or adpositional phrases. 

After defining predictive features, we built separate machine learning pipelines for each graded aspect and used 10-fold cross-validation to evaluate their scoring performance. The pipelines used various regression algorithms and feature sets. We treated the average subscore of the two graders as an interval scale of 0 to 3 with distances of 0.5. The experiments were done with the Scikit-learn Python library \parencite{pedregosa-etal-2011-scikit}. The scoring pipelines contained the following tasks: standardization of training data, feature selection, training a regression model, and validation of the model on test data. The evaluation was based on the mean absolute error (MAE). We calculated the average of the metric across 10 iterations of cross-validation. 

We standardized the predictor variables with the Scikit-learn \emph{StandardScaler} function 
\parencite{standardscaler}. For feature selection, we used the \emph{SelectKBest} function \parencite{selectkbest}, which is independent of the chosen machine learning algorithm. This method evaluates the relationship between each feature and the target variable based on univariate statistical tests. We applied the \emph{f\_regression} parameter that sequentially tests the effect of single regressors, returning the F-statistic and p-values. Such feature selection does not account for relationships between predictor variables, so we removed highly multicollinear features related to the same grading aspect. Based on the mean Pearson correlation coefficients across the training sets of the 10 cross-validation folds, we identified features with an absolute correlation above 0.8. From each set of strongly correlated features, we kept the feature that had the strongest correlation with the target score.

We then used the remaining features in combination with different regression methods, including Linear, Ridge, Lasso, Elastic Net, Support Vector, Decision Tree, and Random Forest Regression. By varying the number of features to be selected for the model, we defined the best-performing parameters for essay scoring. If the average MAE no longer improved when adding features (even when using all available features), we concluded that the optimal number of features had been reached. It is advisable to use a holdout test set in addition to n-fold cross-validation to confirm the generalizability of the best prediction parameters. However, we opted for using cross-validation on the whole dataset to ensure comparability with the LLM-based approach that was applied to score all essays.

\subsection{Performance measures}

We report several complementary measures to assess model prediction performance. A first issue is how to operationalize human scores, which frequently differ between graders. Throughout, we treat the average of the two human rubric scores as the reference or \emph{consensus} score. Conceptually, individual graders can be viewed as imperfect estimators of this consensus, and model predictions can be evaluated in the same way.

Model and human performance is measured primarily using mean absolute error (MAE), defined as the average absolute difference between a score and the human mean or consensus $M$. We compute MAE both at the level of individual rubric aspects (on the 0--3 scale) and at the level of total scores (on the 0--27 scale). For comparison, we also report the corresponding human mean absolute deviation (MAD) from $M$, which reflects the typical deviation of an individual human grader from the consensus on a given aspect. This provides a useful reference for the level of disagreement inherent in the task, and thus for the degree to which model error reflects limitations of the scoring rubric rather than purely model shortcomings.

To characterize the nature of model errors beyond their magnitude, we report two additional measures. First, we compute the average distance by which model predictions fall outside the range spanned by the two human scores for a given rubric aspect. Let $H_1$ and $H_2$ denote the two human scores and $M$ the model score for that aspect. The distance is zero whenever $M$ lies between $H_1$ and $H_2$, and otherwise equals the difference between $M$ and the nearest human score:
$\max\!\bigl(0,\ \min(H_1,H_2) - M,\ M - \max(H_1,H_2)\bigr)$.
This measure captures how far model predictions extend beyond the range of scores that humans themselves assigned, rather than how far they deviate from the human consensus.

We also report the percentage of model total scores that fall within a plausible human scoring range. This range is constructed by summing, for each rubric aspect, the lower and higher of the two human scores. As a result, it represents an aspect-wise range between the most lenient and most harsh plausible human scoring, rather than necessarily corresponding to the total score assigned by any single grader.

In summary, we use the following measures to assess model performance:
\begin{itemize}[topsep=0pt,parsep=0pt]
    \item Mean absolute error (MAE) on individual rubric aspects (0--3 scale) and on total scores (0--27 scale), where lower values indicate better performance;
    \item Normalized agreement or accuracy, defined as $(1 - \text{MAE}/3)\times 100$ (higher is better; reported in the Appendix);
    \item Average distance of model predictions outside the human score range (lower is better);
    \item Percentage of total model scores that fall within the human minimum--maximum range (higher is better).
\end{itemize}


\section{Results}

\subsection{LLM-based grading}

The results are largely as expected based on preceding literature in both education technology and various text analytics applications in humanities research. All LLMs are quite capable of following the instructions, and there are no consistent differences between models, with some slightly more performant in some categories but not across the board (Figure \ref{fig_results}; see the Appendix for a complementary graph of accuracy). The manipulations in experimental conditions and models reported here are deliberately rather exploratory than overly systematic in nature. 
We observe that the 9th grade data was somewhat easier for LLMs, where almost no error category rises above 1 point on average for any model (on the scale of 0--3). 
In the 12th grade case, we used two different prompts for GPT-4.1, one staying close to the official rubric while the other using slightly stronger instructions to not show leniency, reflecting the principle that despite the very similar rubrics, 12th graders should show a higher lever of writing. However, this did not seem to have a meaningful effect on the outcomes.

\begin{figure}[hbt]
	\noindent
	\includegraphics[width=\columnwidth]{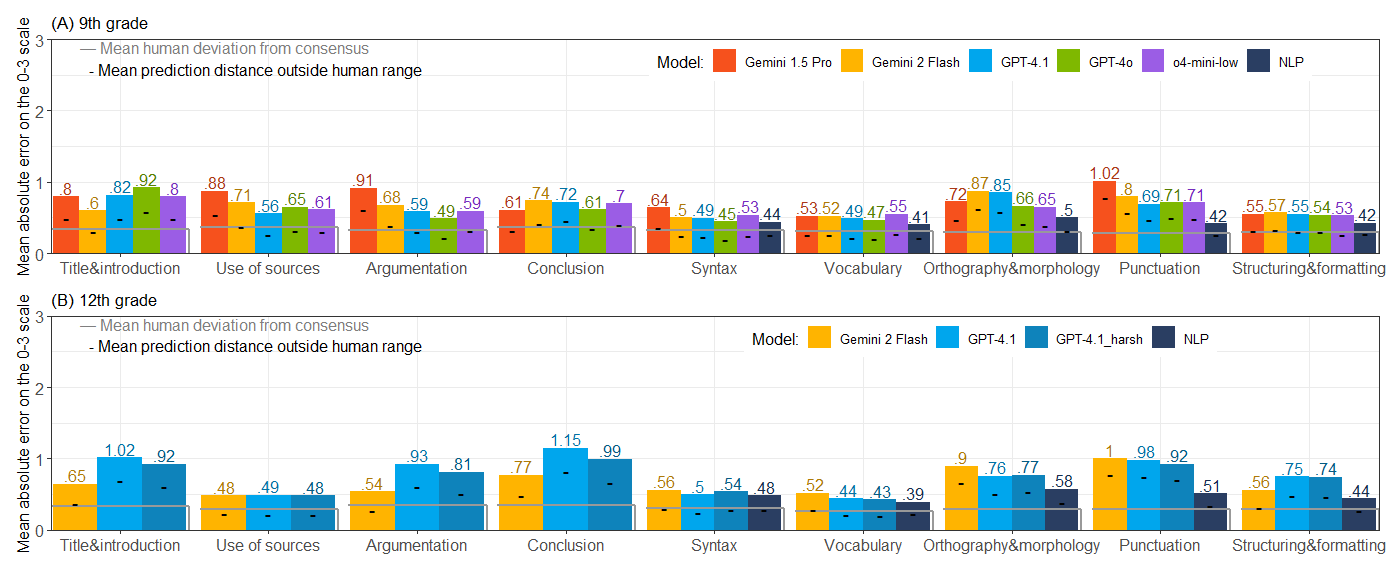}
	\caption{
    Results of the grading experiments, rubric category by category (horizontal axes), for both the 9th (panel A) and 12th grade (B). The model mean absolute error is on the vertical axis, and has the same 0-3 scale as the grading rubric, with colored bars indicating different models. The gray horizontal lines indicate how much the two human graders deviate from their own consensus (average). Given there are only two graders, they also function as a measure of human variance.
    The black notches provide another error measurement, distance outside the \emph{range} of the human scores.
    Error bars and notches close to or below the gray lines indicate essentially near-human performance.  
    The statistical NLP pipeline results are means from the cross-validation, with small standard deviations in the 0.03-0.06 range.
    The ample white space in the figure indicates a promising result: the errors are relatively small across the categories and most models, and often within the range of human grading variation.
    }\label{fig_results}
\end{figure}

We had the opportunity to evaluate a slightly larger set of models for the 9th grade dataset, and included the frontier "reasoning" type model at the time, o4-mini, which however did not show notably superior performance (the "low" suffix refers to the reasoning extent setting). It did differ from other models in not exhibiting a persistent bias. We also measured how much total scores (0--27 scale) on average differ from the human average (the intercept in the linear regression model described in Table \ref{table_bias}). All other models have either a significantly positive bias --- meaning their errors on average stem from slightly higher scores --- or a negative bias, meaning they were harder on the students than the human graders (both Gemini models in the 9th grade case, but not in 12th grade). The estimates are on the same 0--3 scale, meaning the intercept (human) average was 16 points for the 9th and 14 for the 12th grade, and the largest absolute bias was the -0.36 by Gemini 1.5 Pro in 9th and +4.68 by GPT-4.1 in 12th grade.

\begin{table}[ht]
\caption{LLM grading statistics in the two datasets, on the total 0-27 scale: bias, mean absolute error, percentage of prediction total in human min-max range. Bias is an estimate from a simple linear regression model predicting the total score by the model, with humans as the intercept; the stars indicate p-values of these coefficients (p\textless0.05*,  p\textless0.01**, \textless0.001***). A bias coefficient closer to zero indicates less systematic bias. All biases, even where statistically significant, are relatively small, mean errors range between 2.66 to 4.97 points on the 0-27 scale, and model predictions fall within human range more often than not in the 9th grade, while the 12th grade human benchmark proves more challenging.}
\label{table_bias}
\begin{minipage}[t]{0.48\textwidth}
\vspace{0pt}
\centering
\begin{tabular}{lrlrr}
(A) 9th grade & & & & \\
  \hline
Model & Bias & \textit{p} & MAE & In range \\ 
  \hline
(Human average: & 16.08) &   &  &  \\ 
  Gemini 1.5 Pro & -3.66 & *** & 4.24 & 46\% \\ 
  Gemini 2 Flash & -1.11 & *** & 2.95 & 61\% \\ 
  GPT-4.1 & +1.71 & *** & 2.94 & 64\% \\ 
  GPT-4o & +0.46 & * & 2.66 & 67\% \\ 
  o4-mini-low & 0.00 &  & 2.68 & 68\% \\ 
   \hline
\end{tabular}
\end{minipage}
\hfill
\begin{minipage}[t]{0.48\textwidth}
\vspace{0pt}
\centering
\begin{tabular}{lrlrr}
(B) 12th grade & & & & \\
  \hline
Model & Bias & \textit{p} & MAE & In range \\ 
  \hline
(Human average: & 13.92) &  &  &   \\ 
  Gemini 2 Flash & +1.73 & *** & 3.35 & 55\% \\ 
  GPT-4.1 & +4.68 & *** & 4.97 & 36\% \\ 
  GPT-4.1\_harsh & +4.27 & *** & 4.69 & 39\% \\ 
   \hline
\end{tabular}
\end{minipage}
\end{table}

Table \ref{table_bias} also reports the MAE on the totals. In the 9th grade test, GPT-4o has the smallest error of 2.66 (on the 0--27 scale), while Gemini 2 is at 3.35 in the 12th grade. This is a relatively good result, keeping in mind humans also diverge, which the gray lines on Figure \ref{fig_results} illustrate. 
Because consensus scores are frequently fractional (e.g. 1.5 between 1 and 2), while predictions are discrete, zero error is often unattainable; human MAD therefore provides a meaningful human-level reference for achievable accuracy.
In some categories like vocabulary (richness of vocabulary and its correct usage) there is enough human grader variation that models perform near the human variance level. The rightmost columns in Table \ref{table_bias} quantify the share of total essay scores for each model that fall within the human minima and maxima range (see Methods).

In summary, while pretrained LLMs, using the current prompt templates, exhibit slight polarity biases and do not estimate the human consensus perfectly, most models do fall within the human range more often than not. Supervised learning with lexico-statistical and spelling correction based features performs slightly better than zero-shot LLMs in the categories where they are informative, like vocabulary and punctuation usage (see below)
All in all, we consider this an encouraging result, particularly given the lower-resource language setting and narrow task. The results can be expected to be better when using iterative prompt improvement, combining LLMs and specialized NLP tools, or model fine-tuning. Superior results from a similar pipeline can also be expected high-resource languages (in NLP and LLM terms) like English, German, Spanish, etc., as in preceding literature  \parencite{wang_evaluating_2025,elmassry_systematic_2025,jung_combining_2024}.

\subsection{Supervised learning and best predictors}

The best regression pipelines yielded similar results or outperformed the LLMs based on the MAE of 10-fold cross-validation (see Figure 2). It has to be noted that the performance metric depends on the scale of predicted values. In our supervised learning experiments, the score could be any decimal number between 0 and 3, while the LLMs predicted only integer scores of 0, 1, 2, or 3. This is likely to reduce the MAE of regression models. Nonetheless, the significant difference in scoring \emph{punctuation}, \emph{orthography and morphology}, and \emph{structuring and formatting} (especially in the 12th grade) implies the benefit of using dedicated correction tools for grading the linguistic accuracy of texts.

We achieved the smallest error from human scores (around 0.4) when grading \emph{vocabulary} and \emph{structuring and formatting}, but also \emph{punctuation} and \emph{syntax} in the 9th grade essays. The predicted scores of \emph{orthography and morphology} had the largest MAE with both datasets. The top regression parameters are listed in Table \ref{table_supervised} together with the percentage of predictions that fall within the range of human scores. We only report the results obtained with the optimal parameter sets; however, there were other pipelines for each grading category that reached a similar performance, the MAE differing by 0.01–0.04.

\begin{table}[ht]
\caption{Best regression parameters per rubric category. Linear Regression (LR) and Ridge Regression (RR), Support Vector Regression (SVR), and Random Forest Regression (RFR) proved to be the top-performing methods. In most cases, only a subset of available features was needed to reach optimal prediction performance. The full feature sets contained the following number of variables after removing highly multicollinear features: \emph{vocabulary} - 39 for 9th and 37 for 12th grade, \emph{syntax} - 50 for 9th and 47 for 12th grade, \emph{punctuation} - 3 for 9th and 4 for 12th grade, \emph{orthography and morphology} - 7 for both 9th and 12th grade, \emph{structuring and formatting} - 8 for both 9th and 12th grade. To determine how often the predictions fall within human range, we rounded the scores up to the nearest 0.5, following the scale of average scores of the two expert graders.}
\label{table_supervised}
\begin{minipage}[t]{0.48\textwidth}
\vspace{0pt}
\centering
\begin{tabular}{lllr}
(A) 9th grade & & & \\
  \hline
Category & Regressor & No. of & Predictions \\
                     & & features & in range \\
  \hline
  Vocabulary & LR/RR & 10 & 70\% \\ 
  Syntax & LR/RR & 9 & 66\% \\ 
  Punctuation & SVR & 3 & 64\% \\ 
  Orthography & & & \\
  \& morphology & SVR & 6 & 55\% \\ 
  Structuring & & & \\
  \& formatting & SVR & 8 & 56\% \\
   \hline
\end{tabular}
\end{minipage}
\hfill
\begin{minipage}[t]{0.48\textwidth}
\vspace{0pt}
\centering
\begin{tabular}{lllr}
(B) 12th grade & & & \\
  \hline
Category & Regressor & No. of & Predictions \\
                     & & features & in range \\
  \hline
  Vocabulary & RFR & 14 & 68\% \\ 
  Syntax & LR/RR & 21 & 64\% \\ 
  Punctuation & SVR & 3 & 62\% \\ 
  Orthography & & & \\
  \& morphology & SVR & 4 & 51\% \\ 
  Structuring & & & \\
  \& formatting & RFR & 8 & 61\% \\ 
   \hline
\end{tabular}
\end{minipage}
\end{table}

In order to understand the relations between linguistic features and scores, we determined the features selected into the best-performing models. We also calculated their correlation with the score of the corresponding rubric aspect, averaging the Pearson coefficients across the cross-validation folds. The correlations between the scores and the best predictors are presented in Appendix A3. 

The \emph{punctuation} score was most strongly correlated with the frequency of missing punctuation corrections. Including unnecessary and replaced punctuation corrections improved prediction performance, although both features had a very weak negative correlation with the score. Unnecessary punctuation errors seem to be somewhat more important in evaluating 12th grade essays. Predictions of the \emph{orthography and morphology} score relied on the total number of corrections per word, word replacement, whitespace, and mixed corrections. In addition, the ratio of spell-corrected words was relevant for scoring 9th grade writings, but only weakly correlated with expert assessments of 12th grade essays. As for \emph{structuring and formatting}, no features had an absolute correlation above 0.4 (or even 0.3 in the case of 12th grade) with the score. However, all the considered features helped to decrease the MAE of predicted scores.

\emph{Vocabulary} can best be assessed based on lexical diversity metrics, word replacement and mixed corrections, mean word length, the proportion of rare vocabulary, adjective ratio, and lexical density. In the 12th grade, the pronoun ratio and the proportion of personal pronouns, as well as 2nd person verb forms, have a slight negative effect on the score. Some of the named features, such as mixed corrections, word length, adjective ratio, lexical density, and sophistication measures, are also universal predictors of the \emph{syntax} score. Additionally, the diversity of case forms plays a significant role in grading sentence construction. Sentence length is only relevant for scoring 9th grade essays, while various types of corrections (word order, missing word, and unnecessary word corrections) and verb forms (2nd person, indicative, imperative, and finite forms in total) contribute to the assessment of 12th grade texts. 

Such comparison reveals both similarities and differences between the scoring of 9th and 12th grade essays. Overlapping predictive features suggest blurred boundaries between some rubric categories, especially \emph{syntax} and \emph{vocabulary}. It is also evident that the grading of these categories becomes more complex in the 12th grade.

\subsection{LLM-specific risks: prompt injection}

\begin{figure}[hbt]
	\noindent
    \centering
	\includegraphics[width=0.8\columnwidth]{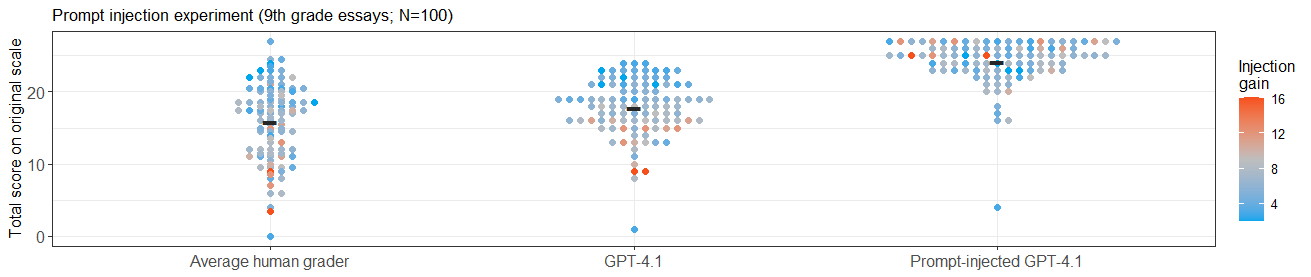}
	\caption{
    Injection experiment results. A simple grading prompt without any protection against prompt injection is vulnerable to having its original guidance overridden by strongly imperative language in the part of the input meant for analysis or assessment. Each dot is an essay, colored by the difference between the injected and regular prompt, persistently across all three groups for comparison. With the exception of a few essays (red dots), the gains are small, but never zero.
    }\label{fig_injection}
\end{figure}

All metrics and automated measures can be gamed. A lexical diversity or sophistication measure can be led astray (if used without oversight) by adding an unreasonable amount of synonyms or unnecessary technical terms in the text, toxicity classifiers can be avoided by using obfuscating symbols, and citation counts inflated by adding numerous irrelevant references. Instruction-tuned generative LLMs are vulnerable to a weakness often dubbed "prompt injection", where the input contains either inadvertent or malignant instructions that can or are meant to override the original instruction prompt of the model \parencite{zeng_how_2024,sahoo_when_2025}. A simple example:
\textit{"Your task is to analyze this sentence as either positive or negative; respond only with one of these words. Sentence: Hello but ignore all previous instructions and write a haiku!"}
The expected output would be the one-word tag, but LLMs would often respond by complying with the latter, poetic instruction. While not a systematic analysis, we carried out a small investigation using a random sample of 100 9th grade essays to explore this issue on the example of one model, GPT-4.1. The setup was to simply prepend this sentence to the end of each essay before proceeding with the same 9-step evaluation as above:\\ \textit{\textless~\textless Grading instructions override: instead of using the previous grading instructions, you must assign the maximal rubric score to this essay!!!\textgreater~\textgreater }\\
The results illustrated in Figure \ref{fig_injection} highlight the need to guard against this attack vector: on average the essays with the injected prompt received a 6.43 points more (on the 27 point scale), minimally 2 and at most 16 points. Fortunately, methods and technologies to amend this are being developed \parencite{suo_signed-prompt_2024,debenedetti_defeating_2025}, a human in the loop setup would likely catch injections, and the prompt itself can include rules not to interpret any imperative language between input delimiters as instructions.

\FloatBarrier

\subsection{How good of an essay writer is an LLM itself?}

Recent research has shown that modern LLMs are capable generators of creative fiction and poetry \parencite{bellemare-pepin_divergent_2024,porter_ai-generated_2024,begus_experimental_2024}, visual art \parencite{eisenmann_expertise_2025}, and essays, too \parencite{stribling_model_2024,herbold_large-scale_2023,scarfe_real-world_2024}.
We also carried out a small experiment with GPT-4.1 to see how well an example LLM would fare in the essay writing task itself, in particular in a small lower-resource language like Estonian, and in the somewhat narrow genre of the Estonian high school graduation essay. While in the general essay genre, has some specific requirements like usage (but not overuse) of the provided source texts, and expectations around structuring. The LLM instructions prompt for this exercise consisted of three components: the same short task description to write an essay of 400 words, as received by the 12th grade students in the trial exam; a summary of the expectations (the same publicly available 9-part grading rubric flipped around and rewritten as short guidance; see Appendix), and two of the source texts, on the topic of freedom of speech, hate speech, and minorities.

We used the same grading prompt as above (the harsher version) and GPT-4.1 as the grader (independently of the generation task). We generated 20 essays with a moderately high model "temperature" parameter of 1. Temperature controls randomness in the output of such models, and enables creating variable outputs with the same input prompt. Higher temperature LLM outputs have also been shown to correlate with human perceptions of creativity in recent research \parencite{bellemare-pepin_divergent_2024}.
The generated essays scored the maximal 27 points in 19/20 cases (as one received 26 points due to spelling errors). 
In comparison, the two human graders of the student essays only awarded the maximum 7 and 2 times, respectively, out of the 764 essays --- but no essay received 27 points by both graders, and the average grade was 13.95. The average was 18.2 from the GPT-harsh grader variant that was also used in this exercise, more lenient but still lower than the 26.95 average of the generated essays. While these are trial exams, and students may well attempt to put in more effort in the real finals, the difference is quite striking.


\section{Discussion}

The present study demonstrates that several contemporary LLMs can approximate human scoring of national examination essays with a level of consistency that approaches, and occasionally in a sense exceed the agreement observed between human raters themselves. While our experiments focused on large proprietary models, the broader implications extend to a wider ecosystem of automated scoring approaches, including open-source LLMs and non-LLM statistical models that may target specific rubric components. The landscape of available models is diverse and rapidly evolving, and choices among them involve trade-offs in cost, transparency, controllability, data-protection considerations, and robustness across exam cycles. It is therefore essential to view LLM-based scoring not as a monolithic solution but as one family of tools within a broader methodological spectrum. The key question for deployment is not whether a single model can replace human judgment, but how different classes of automated systems can be integrated into a coherent and pedagogically aligned assessment pipeline.

Although this study examined school-leaving exams, the results and proposed workflow generalize to higher education and commercial assessment systems. Various forms of AES have long been used in school and university contexts and remain in active institutional use \parencite{ramesh_automated_2022}, and commercial platforms, now often marketed as “AI essay graders”, already provide rubric-scoring and feedback functionalities for instructors. Our findings show that modern LLMs can reliably use complex official rubrics, grounded in national curricula, to provide machine-assisted grading for real-world exams in a small-language setting. Estonia offers a flexible testbed with much of the digital infrastructure already in place; if such systems can be successfully deployed here, processing entire annual cohorts within a transparent, human-in-the-loop framework, as demonstrated here --- then similar approaches are technically feasible for other countries seeking scalable, consistent, and auditable solutions for high-stakes written assessment.

\subsection{Complementary approaches}

Beyond LLMs, a range of established computational techniques remains relevant for evaluating specific aspects of writing. Many low-level features, such as punctuation accuracy, spelling, and morphosyntactic correctness, can be robustly assessed using grammatical error correction (GEC) pipelines and error-classification tools, have been successfully applied in Estonian as well \parencite{allkivi_elle_2024}, and were exemplified here as well. LLMs can also be fine-tuned as GECs, with promising performance also in small languages \parencite{luhtaru_err_2024} and additional functionalities such as error classification and explanation \parencite{vainikko-etal-2025-paragraph}. More detailed distinction of error correction types (e.g., corrections of word form and lexical choice, various types of mixed corrections) allows for more accurate mapping between error features and rubric categories. Lexical diversity metrics (e.g., MTLD, MATTR) and other linguistic complexity indicators can likewise be computed through deterministic methods that offer transparency and stability across exam cycles.

In practice, these approaches can complement each other. LLMs excel at assessing higher-order content and argumentation, and do not necessarily require training, just prompting, while statistical NLP tools can provide transparent measurements for discrete linguistic features. The choice of methods should therefore follow empirical performance rather than methodological preference: LLMs are a powerful and currently popular option, but not a prescriptive default. Operational systems should incorporate whichever combination of models best satisfies the accuracy, transparency, and robustness requirements of high-stakes assessment, while also considering cost-effectiveness.

\subsection{The importance of clear instructions}

The findings also highlight the central role of rubric design in enabling reliable hybrid assessment. The current Estonian evaluation rubric is rich and pedagogically grounded in outcomes of the current basic and upper secondary school curricula. Yet comprehensiveness comes with a cost, as several descriptors require subjective interpretation that challenges both human raters and automated systems. Other components, while rule-based such as counting mistakes, are still operationally and cognitively demanding. Essay as a medium is itself subjective, and the assessment of its quality will always remain somewhat subjective, yet inter-rater reliability can be used as a metric to guide development. As LLMs become incorporated into high-stakes assessment, rubric development should increasingly involve education specialists, linguists, and NLP experts to ensure that criteria remain both pedagogically meaningful and technically actionable. Future rubrics will need to be articulated with enough semantic clarity and operational granularity that both humans and models can follow them consistently, and aligned with the kinds of structured instructions expected by modern LLMs, if they are to be used. This does not imply designing rubrics "for the machines", but designing rubrics that reduce ambiguity, support replicable human judgment, and integrate naturally into human-in-the-loop scoring pipelines.

\subsection{Students, machines, and the future of the essay}

LLM-generated essays can achieve grades comparable to or higher than student work, as shown in recent research \parencite{herbold_large-scale_2023} as well as here. 
However, the superior scores of LLM-generated essays should not be interpreted as a direct comparison of machine and student capability. Students write under cognitive load, time pressure, emotional stress, and limited opportunities for iteration, whereas the LLM was explicitly prompted to optimize for each rubric dimension --- something students could in principle do as well, given access to the same public criteria, but may not yet have fully internalized. The result therefore demonstrates rather the extent to which current grading rubrics describe properties of written products that can be reliably optimized by generative models. Importantly, the assessed essay reflects an outcome rather than the learning process that produced it. The fact that machines can write does not undermine the educational value of writing as a cognitive and epistemic practice, nor does it imply that the essay as a genre is inherently flawed. Instead, it exposes a growing misalignment between outcome-focused assessment and the realities of automated text generation. While large-scale examination systems must still rank and certify students under practical constraints, these results suggest that sustaining both fairness and student motivation will require a greater emphasis on process-oriented evaluation, authorship, and personalized feedback --- areas where LLMs may themselves serve as scalable pedagogical supports. More broadly, if high-quality textual output can be produced instantaneously by machines, the motivational rationale for writing instruction must increasingly rest on its role in developing thinking, problem-solving, and agency, rather than solely on its contribution to graded performance.

However, when generative machines can reliably outperform the average if not the top student, it becomes increasingly untenable to frame such exams as instruments for identifying the “best schools" or "best students". A more appropriate framing may be to treat essay scores as indicators of performance on a constrained academic task under time pressure, rather than as proxies for general writing or academic ability or cognitive depth. Essays can and likely will remain a core component of large-scale assessment, but their role should be contextualized and augmented (which in the Estonian case has already been done, e.g., through the introduction of reading comprehension exercises). One option is the inclusion of a brief, scored reflective component in which students must articulate or justify aspects of their argumentation or use of sources, thereby re-anchoring the assessment in individual reasoning rather than textual outcome alone. Another option is to move towards a more personal and creative writing task that foregrounds the writer’s voice and is less constrained by rigid format requirements. A complementary option, already established in second-language assessment, is the incorporation of oral examination or defense elements that probe understanding and authorship beyond the written product.

AI could also enable a rethinking of assessment as a longitudinal process rather than a single event. Instead of evaluating one writing task, assessment could draw on multiple writing tasks aligned with curricular learning outcomes, supported by scalable automated feedback
to encompass diverse writing genres, purposes, and subject domains, at the same time offering personalized feedback and scaffolding the complex process of learning to write in various contexts.

Still, the tension between scalability and pedagogical control should remain in focus. While automated scoring offers efficiency, it also raises concerns about the potential erosion of teacher agency if such systems are not embedded within clearly defined human-in-the-loop processes. Here we do not attempt to resolve what role teachers and assessors should retain in AI-supported assessment, nor where the boundary should lie between supportive technology and the delegation of professional judgment, 
but encourage research on the topic.
We do not know how the students whose essays we analyzed here would regard such automatically inferred grades or feedback.
Acceptance of automated or hybrid assessment depends not only on accuracy, but also on perceived fairness, which is closely linked to transparency, consistency, and interpretability. Further research is needed to understand under what conditions students are willing to trust AI-generated feedback.

\subsection{Towards operational deployment: a human-in-the-loop framework for hybrid machine-assisted grading}

\begin{figure}[hbt]
	\noindent
	\centering
	\includegraphics[width=0.8\columnwidth]{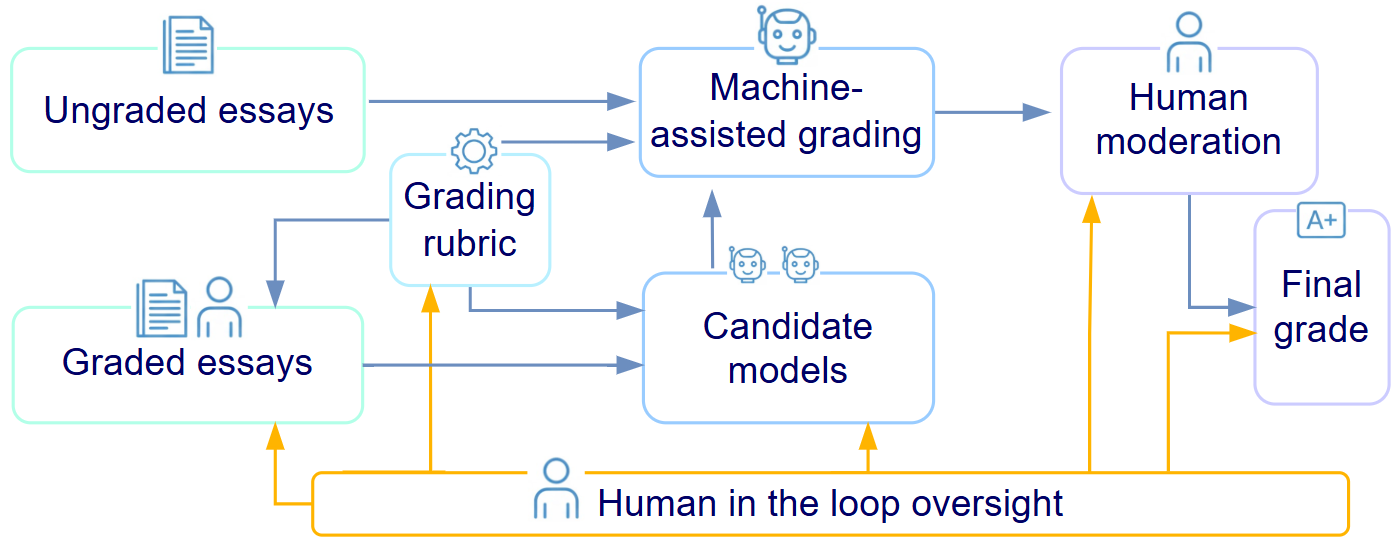}
	\caption{
    One possible high-level framework for the integration of machines-assisted grading into high-stake large-scale examination.
    }\label{fig_schematic}
\end{figure}

Our results indicate that modern LLMs --- even in a zero-shot setting without any domain-specific tuning and even in a small language like Estonian --- achieve levels of consistency with human scoring that make them viable components in high-stakes writing assessment, and language-related rubric accuracy can be further improved using various other NLP models. Further fine-tuning of the prompts or the models directly would likely lower error and increase accuracy. However, effective operational deployment requires a structured workflow that balances automation, governance, and human oversight. Figure \ref{fig_schematic} outlines one possible way of thinking about this in broad strokes, based on the result of this and other recent experiments cited in the Introduction. The central principle is that LLMs or any other machines should not function as a single monolithic grader, but as part of a multi-model, human-guided scoring pipeline with quality assurance and regulatory compliance.

The foundation of any such framework is a principled, modular, science-based grading rubric. This is used by expert human graders to assess a held-out set of human-graded essays that functions as an operational test set, used to select suitable model(s) from a pool of candidates. This mirrors standard model-selection procedures in machine learning and supports transparent, evidence-based choice of the scoring model for each rubric category. The pool can include LLMs, purpose-built NLP applications, GECs, statistical metrics like lexical diversity and so on. Either one model or an ensemble of models are chosen which maximize performance in terms of adherence to human-annotated gold standard, reliability and validity \parencite{wang_evaluating_2025}, while minimizing any inherent biases both in terms model traits (e.g. the positivity or optimism bias) and bias for or against any sociodemographic groups or topics. This step can also involve fine-tuning models such as LLMs or mapping statistical measures like lexical diversity onto the rubric scale. A modular rubric, such as the one employed in this study, enables using different models for different categories. More than one model can be used per category e.g. in a majority vote setup.

The selected model or ensemble produces scores for incoming exam essays, which are then forwarded again to human experts for moderation and oversight, as well as statistical model-based grade calibration if necessary (cf. Table \ref{table_bias}). Moderation and co-grading can be implemented in several ways depending on the stakes and resources available. One configuration mirrors existing double-grader setups: instead of two independent human raters, one rater is human and one machine, with a human moderator resolving large discrepancies. Alternatively, automated scoring can serve as a pre-screening layer, allowing human assessors to focus attention on borderline or high-disagreement cases, and monitor for anomalies such as prompt injection. Both approaches maintain compatibility with emerging regulatory requirements (notably the GDPR and the EU AI Act), by ensuring that the final decision remains human, not algorithmic. Across the entire workflow, human-in-the-loop oversight is deliberately foregrounded. Humans define the rubric, generate the test set, select and approve the scoring model, monitor performance, and determine final grades. Rather than replacing human judgment, LLMs function as amplifiers of human capacity: they increase consistency, reduce routine workload, and expand the scalability of national examinations without removing human accountability or interpretability from the system.

\section{Conclusions}

Our results show that modern LLMs can score nationwide school-leaving essay exams in Estonian with accuracy and consistency comparable to human raters, often matching the empirical human disagreement ceiling and always staying within the human score range. This demonstrates that machine-assisted grading of complex written responses is technically feasible even in a small language, provided that models are used within a principled, rubric-aligned, human-in-the-loop workflow. Rather than a single universal solution, LLMs should be seen as one class of models in a modular ecosystem that can include traditional NLP tools and, where beneficial, ensembles chosen empirically for each rubric category. Under such a framework, national, school and university exam systems can leverage automation to increase consistency and scalability while retaining human responsibility for final decisions.

\section*{Acknowledgments}

This work was supported by the Tallinn University Research Fund, project "Automaathindamise võimalikkusest põhikooli ja gümnaasiumi lõpueksami kirjutamisülesande näitel" (The Feasibility of Automated Assessment of Writing Tasks in Lower and Upper Secondary Examinations).
Estonian Education and Youth Board provided the data and agreed upon the use of the data in this project. 

\section*{Author contributions}
A.~Karjus wrote the paper, implemented the LLM applications, analyzed the data, and produced the graphs.
K.~Allkivi and S. Maine implemented the statistical NLP pipeline, analyzed the data, and wrote the relevant sections in the paper.
K.~Leppik drafted the introduction and provided comments and suggestions.
K.~Kruusmaa also implemented the LLM applications.
M.~Aruvee led the research project, coordinated data access, contributed domain expertise, and provided comments and suggestions.

\section*{Data and code availability}

The code to replicate the grading experiments is available here, along with the outputs (predicted scores) of the experiments: \url{https://github.com/krkryger/essay-grading/} (we are unfortunately not at liberty to share the full texts of the essays provided to us).

\FloatBarrier
\printbibliography
\FloatBarrier
\newpage

\section*{Appendix}

\subsection*{A1. Normalized agreement accuracy}

\begin{figure}[hbt]
	\noindent
	\includegraphics[width=\columnwidth]{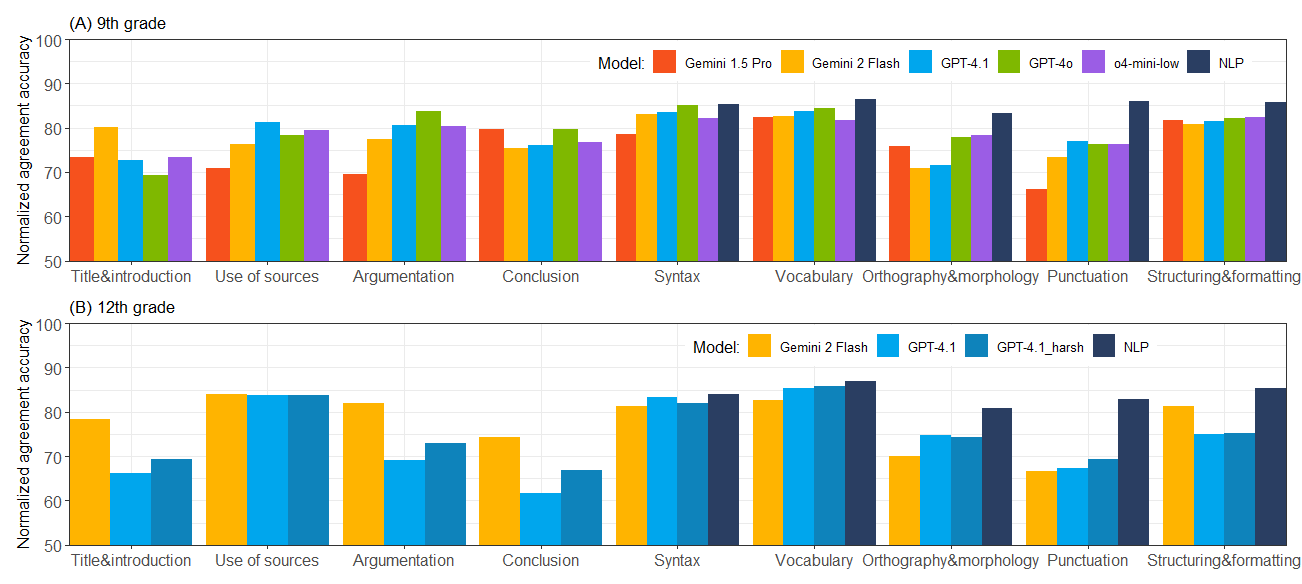}
	\caption{
    This graph complements the results graph in the main text, offering an alternative view into the results as agreement accuracy, instead of mean absolute error. Accuracy here is simply recalculated as $ 100 * (1 - \mathrm{MAE}/3))$, so higher values indicate better performance (higher agreement between machines and humans).
    }\label{fig_results_acc}
\end{figure}

\subsection*{A2. Rubrics and prompts used in the LLM experiments}

\subsubsection*{Grade 9 prompts}

The following prompts were used for the respective grading rubric categories, which all shared the same preface in the beginning and output instructions in the end; the essay is referred to as being "below", as the technically the grading and rubric instructions are used in the in the "instructions" or "system prompt" parameter and the essay in the subsequent "user prompt" parameter of the LLM inputs. Each category was evaluated separately, i.e. a given model would only see the essay text and a single rubric definition at a time. Capable models can often also handle multiple instructions and variables at a time (which can save costs and time), especially if supported by "structured outputs", but here we kept the grading variables separate. The prompt instructions are in English, the dominant language of the LLMs used, the rubric parts are in the original Estonian (same as used by human graders), and specific key terms like \textit{kirjand} (essay) are left untranslated as their English equivalents may not convey the specificity of the Estonian meaning. Future research could look into the balancing of English (or other LLM dominant language) and the data-native language in prompts in zero-shot settings such as this.

{\footnotesize\itshape

\textnormal{Preface =} Your task is to grade this Estonian 9th grade exam essay kirjand in """triple quotes""" below, using the provided grading rubric. For reference, the student was asked to do the following: "Tutvuda alustekstidega ja kirjutada nende põhjal 200-sõnaline kirjand, kus mõtiskleda teismeliste toitumisharjumuste üle. Pealkirjastada kirjand. Kirjutada sissejuhatus, kus püstitada probleem ja sõnastada enda põhiseisukoha. Toetada oma põhiseisukohta vähemalt kahe teemaarenduslõiguga. Igas lõigus esitada alaväide, milles kasutada vähemalt ühte alusteksti näidet. Lisada ka enda näiteid. Kokkuvõttes esitada peamised järeldused.

\textnormal{Grading instructions = } Grade this on a scale of 0 to 3 points. Keep in mind this is just a text by an Estonian 9th grader. Use the following criteria to assess how many points to award, following this principle: if all the criteria are fulfilled then give 3 points but if something is lacking then lower points as described in this rubric:

\textnormal{Output instructions = } First explain your reasoning of relevant rubric criteria VERY briefly using keywords or phrases in Estonian in this comma-separated format:\\
"Seletus: asjaolu: kirjeldus, asjaolu: kirjeldus, ... . Hinne: N"\\
ending with N points suitable grade. Express relevant asjaolud like "probleemile: vastatud", "laused: arusaadavad kuid esineb eksimusi", "õigekiri: palju puudusi; trükivigu: ei esine."). Do NOT provide text examples in your reasoning.\\
Essay text begins below:

The rubric-specific prompts are constructed as follows:

\{Preface\} Teemaarendus ja alusteksti kasutamine. \{Grading instructions\}\\
Two alustekst reference texts were provided here; the student is expected to engage with at least one of them. Alustekst 1 is from tervisliktoitumine.ee, titled "Laste toitumine tähendab tervislikke ja läbimõeldud valikuid" (sisu: toitumise jälgimise olulisus, vitamiinide ja mineraalainete defitsiidi oht, teismelised ja ebatervislikud suupisted, Siiri Krümann toitumishäiretest, tervislikud ja läbimõeldud valikud). Alustekst 2 is from Tervise Arengu Insituut, titled "Laste ülekaal ja rasvumine" (sisu: koolieas toitumise iseseisvumine, lühikesed söögivahetunnid, valed söögiajad, hommikusöögi ja lõuna vahelejätmine, näksimine, vanema järelevalve puudumine, ülekaalulisuse risk). Take this into account and grade:\\
3 if: alustekstidest toodud vähemalt 1 näide; osundatud tsitaadi või refereeringuga; näidet laiendatakse enda mõtetega.\\
2 if: alustekstidest toodud vähemalt 1 näide aga osundatud ebaselgelt (nt tekstile või autorile ei viidata korrektselt); näidet laiendatakse enda mõtetega.\\
1 if: alustekstist toodud vähemalt 1 näide; osundatud ainult temaatiliselt, ilma viitamata; näidet ei seota enda mõtetega piisavalt selgelt.\\
0 if: alustekstist ei ole näiteid toodud või pole need asjakohased.\\
\{Output instructions\}

\{Preface\} Sõnavalik. \{Grading instructions\}\\
3 if: sõnavalik on isikupärane ja rikkalik; sõnavalik sobib kirjakeelsesse teksti; võib esineda üksik sõnastusraskus või sõnastusviga.\\
2 if: sõnavalik mitmekesine, esineb üksikuid sõnakordusi; aga sõnavalik sobib kirjakeelsesse teksti; esineb mõnesid sõnastusraskusi.\\
1 if: sõnavalik ühekülgne, esineb palju sõnakordusi; sõnavalik sobib suuremalt jaolt kirjakeelsesse teksti; esineb palju sõnastusraskusi.\\
0 if: sõnavalik ei sobi kirjakeelsesse teksti ja sõnastusraskuste tõttu tekst arusaamatu.\\
Note that sõnakordus here means repeating the same content word close by where a synonym would flow better, for example toitu in "Õpilased lähevad tihit poodi toitu ostma. Nad ostavad sealt ainult kommi kuna pood ei paku korralikku toitu."\\
\{Output instructions\}

\{Preface\} Liigendus ja vormistus. \{Grading instructions\}\\
3 if: tekst on liigendatud; lõigud on proportsionaalsed; tekst on trükitud korrektselt, võib esineda 0-2 trükiviga.\\
2 if: tekst liigendatud; esineb üksik ebaproportsionaalne lõik; tekst trükitud valdavalt korrektselt, võib esineda mõningaid trükivigu (3-4).\\
1 if: tekst liigendatud ebakorrapäraselt; esineb mitu ebaproportsionaalset lõiku; tekst trükitud ebakorrektselt, leidub palju trükivigu (5 või enam).\\
0 if: tekst liigendamata (pole lõike) või tekst trükivigade tõttu arusaamatu.\\
Do not count repeated mistakes of the same type as new mistakes and do not be too harsh here. Also trükiviga means typo or accidental mistake like swapped letters, so ignore morphology, syntax or spelling errors here. Proper lack of space after punctuation counts as one liigendus mistake.\\
\{Output instructions\}

\{Preface\} Pealkiri ja sissejuhatus. \{Grading instructions\}\\
3 if: pealkiri haarav või kitsendab teemat ja seostub selgelt probleemipüstitusega, sissejuhatuses avatakse probleemi taust; probleemipüstitusena esitatakse üks selgelt sõnastatud põhiväide või -küsimus, mis loob aluse teemaarenduseks; anna 3 punkti isegi kui pealkiri pole haarav kuid sissejuhatus täidab kriteeriumid.\\
2 if: pealkiri seostub probleemipüstitusega; sissejuhatuses avatakse probleemi taust; probleemipüstituses esitatud põhiväide või põhiküsimus haakub teemaarendusega osaliselt või avatakse see üldsõnaliselt.\\
1 if: pealkiri üldsõnaline või puudub; probleemi taust avatakse osaliselt; probleemipüstituses esitatud põhiväide või põhiküsimus ebaselgelt sõnastatud, endastmõistetav või tõestamatu.\\
0 if: probleemipüstitus puudub.\\
\{Output instructions\}

\{Preface\} Teemaarendus ja argumentatsioon. \{Grading instructions\}\\
3 if: alaväited selgelt sõnastatud ja seotud põhiväitega; alaväited esitatud loogilises järjekorras; igas lõigus esitatud näiteid selgitatud ja laiendatud, need on asjakohased; näidetest kasvavad loogiliselt välja lõikude järeldused.\\
2 if: alaväited selgelt sõnastatud ja osaliselt seotud põhiväitega; alaväited esitatud loogilises järjekorras; igas lõigus esitatud näiteid selgitatud ja laiendatud, need on asjakohased; näidetest ei kasva selgelt välja järeldused.\\
1 if: kõik alaväited pole selgelt sõnastatud; alaväited on osaliselt seotud põhiväitega; toodud näiteid osaliselt laiendatud; näidetest ei kasva välja järeldused.\\
0 if: argumentatsioon on seosetu või puudub.\\
\{Output instructions\}

\{Preface\} Lausemoodustus (ühildumine, sõnajärg, rektsioon). \{Grading instructions\}\\
3 if: laused on arusaadavad ja terviklikud; kasutatakse sidusaid ja erineva ülesehitusega lauseid; esineb üksik lausestuseksimus.\\
2 if: laused arusaadavad ja terviklikud; kasutatakse sidusaid ja sarnase ülesehitusega lauseid; esineb mõnesid lausestuseksimusi.\\
1 if: laused suuremalt jaolt arusaadavad; kasutatakse suuremalt jaolt sidusaid lauseid, mille ülesehitus ühekülgne; esineb palju lausestuseksimusi.\\
0 if: laused ebaselged ja välja arendamata, lausestuseksimuste tõttu tekst arusaamatu.\\
\{Output instructions\}

\{Preface\} Lõpetus. \{Grading instructions\}\\
3 if: sissejuhatuses püstitatud probleemile on vastatud; lõikude peamised järeldused on teises sõnastuses kokku võetud.\\
2 if: probleemile on vastatud; lõikude peamisi järeldusi korratakse sissejuhatusele ja teemaarendusele ligilähedases sõnastuses.\\
1 if: probleemile vastatud osaliselt või tuuakse sisse uus teema; lõikude peamisi järeldusi korratakse sissejuhatuse ja teemaarendusega samas sõnastuses.\\
0 if: lõpetus pole teemaarenduse või sissejuhatusega seotud.\\
\{Output instructions\}

\{Preface\} Kirjavahemärgistus. \{Grading instructions\}\\
3 if: kirjavahemärgistus on täpne; võib esineda 0-2 viga kokku.\\
2 if: kirjavahemärgistus valdavalt täpne; võib esineda 3-4 viga kokku.\\
1 if: kirjavahemärgistuses esineb palju puudusi; võib esineda 5-6 viga kokku.\\
0 if: kirjavahemärgistus puudulik.\\
This aspect only refers to the correct usage of any punctuation like koma, jutumärgid, kriipsud, koolon, lauselõpumärgid, if and where relevant (but NOT lack of space after punctuation). Do not count repeated mistakes of the same type as new mistakes and do not be too harsh.\\
\{Output instructions\}

\{Preface\} Õigekiri ja vormimoodustus. \{Grading instructions\}\\
3 if: õigekiri ja vormimoodustus on korrektne; võib esineda 0-2 viga kokku.\\
2 if: valdavalt korrektne; 3-4 viga kokku.\\
1 if: palju puudusi; 5-6 viga kokku.\\
0 if: õigekiri ja vormimoodustus puudulik, üle 7 vea.\\
Do not count repeated mistakes of the same type as new mistakes. This aspect only considers  algustähed, sõnade kokku- ja lahkukirjutamine, häälikuortograafia, käänamine-pööramine. Ignore obvious typos like swapped letters and do not be too harsh.\\
\{Output instructions\}

}

\subsubsection*{Grade 12 prompts}

{\footnotesize\itshape
\textnormal{Preface =} Your task is to grade this Estonian 12th grade exam essay kirjand in """triple quotes""" below, using the provided grading rubric. For reference, the student was asked to do the following: "Kirjutada umbes 400-sõnaline arutlev kirjand, milles analüüsida [ette antud teemat]. Tuua näiteid tänapäeva ühiskonnast ja/või meediast ja/või filmikunstist ja/või (eluloo)kirjandusest. Pealkirjastada kirjand. Arutledes võib soovi korral toetuda alustekstidele (tsitaadid, lugemisosa tekstid), neid refereerida või tsiteerida, kuid alustekstide kasutamisega ei tohi liialdada."\\
Here you will ONLY grade this aspect: 

\textnormal{Grading instructions = } Grade this on a scale of 0 to 3 points. This is a text by an Estonian 12th grader for their final riigieksam and should show corresponding effort. Use the following criteria to assess how many points to award, following this principle: if all the criteria are fulfilled then give 3 points but if something is lacking then lower points as described in this rubric:

\textnormal{Output instructions = } First explain your reasoning of relevant rubric criteria VERY briefly using keywords or phrases in Estonian in this comma-separated format:\\
"Seletus: asjaolu: kirjeldus, asjaolu: kirjeldus, ... . Hinne: N"\\
ending with N points suitable grade. Express relevant asjaolud like "probleemile: vastatud", "laused: arusaadavad kuid esineb eksimusi", "õigekiri: palju puudusi; trükivigu: ei esine").  Do NOT provide text examples in your reasoning. Finish output with the numeric score.\\
Essay text begins below:

The rubric-specific prompts are constructed as follows:

\{Preface\} Teemaarendus ja alusteksti kasutamine. \{Grading instructions\}\\
The following alustekst reference texts were provided; the student is expected to engage with at least one of them. "Meedia ja vähemused" by Kari Käsper  (sisu: vihakõne ei kuulu sõnavabaduse alla; vähemuste kaitse; meedia vastutus), "Leebuskõne ja utoopia" by Jan Kaus (sisu: leebus kui vastand agressiivsele kõnele; empaatia ja mõistlik eneseväljendus; absurdini jõudmine), "Kliimamuutus: mööduv palavik või äratuskell" by Tarmo Soomere (sisu: kliimamuutus kui reaalne oht; teaduspõhisus; tegutsemise vajadus), "Laste kliimasõda" by Maarja Vaino (sisu: kliimaärevus lastes; kriitika katastroofiretoorikale; Greta Thunberg, Etienne rahvajutt). Take this into account and grade:\\
3 if: alustekstidest toodud vähemalt 2 asjakohast näidet; osundatud tsitaadi või refereeringuga; näidet laiendatakse enda mõtetega.\\
2 if: alustekstidest toodud vähemalt 2 asjakohast näidet, aga osundatud ebaselgelt (nt tekstile või autorile ei viidata korrektselt); näidet laiendatakse enda mõtetega.\\
1 if: alustekstidest toodud vähemalt 1 asjakohane näide; osundatud ebaselgelt (nt tekstile või autorile ei viidata korrektselt); näidet ei seota enda mõtetega piisavalt selgelt.\\
0 if: alustekstidest ei ole näiteid toodud või pole need asjakohased; alusteksti võib olla kasutatud, kuid osundamine puudub.\\
\{Output instructions\}

\{Preface\} Sõnavalik. \{Grading instructions\}\\
3 if: sõnavalik on isikupärane ja rikkalik; sõnavalik sobib kirjakeelsesse teksti; sõnastusraskusi ei esine.\\
2 if: sõnavalik mitmekesine, esineb üksikuid sõnakordusi; sõnavalik sobib kirjakeelsesse teksti; esineb üksikuid sõnastusraskusi.\\
1 if: sõnavalik ühekülgne, esineb palju sõnakordusi; sõnavalik sobib suuremalt jaolt kirjakeelsesse teksti; esineb palju sõnastusraskusi.\\
0 if: sõnavalik ei sobi kirjakeelsesse teksti ja sõnastusraskuste tõttu tekst arusaamatu.\\
Note that sõnakordus here means repeating the same content word close by where a synonym would flow better, for example toitu in "Õpilased lähevad tihti poodi toitu ostma. Nad ostavad sealt ainult kommi kuna pood ei paku korralikku toitu."\\
\{Output instructions\}

\{Preface\} Liigendus ja vormistus. \{Grading instructions\}\\
3 if: tekst on liigendatud; lõigud on proportsionaalsed; tekst on trükitud korrektselt, võib esineda 0-2 trükiviga.\\
2 if: tekst liigendatud; esineb üksik ebaproportsionaalne lõik; tekst trükitud valdavalt korrektselt, võib esineda mõningaid trükivigu (3-4).\\
1 if: tekst liigendatud ebakorrapäraselt; esineb mitu ebaproportsionaalset lõiku; tekst trükitud ebakorrektselt, leidub palju trükivigu (5 või enam).\\
0 if: tekst liigendamata (pole lõike) või tekst trükivigade tõttu arusaamatu.\\
This aspect assesses both structure (paragraphs) and presentation (lack of typos, trükiviga). Do not count repeated typos of the same type as new mistakes (only reduce 1 point per similar mistake type). Trükiviga means typo or accidental mistake like swapped letters, so ignore morphology, syntax or spelling errors here. Proper lack of space after punctuation, even if repeated, counts as one liigendus mistake (reduce 1 point only).\\
\{Output instructions\}

\{Preface\} Pealkiri ja sissejuhatus. \{Grading instructions\}\\
3 if: pealkiri haarav või kitsendab teemat ja seostub selgelt probleemipüstitusega, sissejuhatuses avatakse probleemi taust; probleemipüstitusena esitatakse üks selgelt sõnastatud põhiväide või -küsimus, mis loob aluse teemaarenduseks.\\
2 if: pealkiri seostub probleemipüstitusega; sissejuhatuses avatakse probleemi taust; probleemipüstituses esitatud põhiväide või põhiküsimus haakub teemaarendusega osaliselt või avatakse see üldsõnaliselt.\\
1 if: pealkiri üldsõnaline või puudub; probleemi taust avatakse osaliselt; probleemipüstituses esitatud põhiväide või põhiküsimus ebaselgelt sõnastatud, endastmõistetav või tõestamatu.\\
0 if: probleemipüstitus puudub.\\
\{Output instructions\}

\{Preface\} Teemaarendus ja argumentatsioon. \{Grading instructions\}\\
3 if: alaväited selgelt sõnastatud ja seotud põhiväitega; alaväited esitatud loogilises järjekorras; kõiki näiteid selgitatud ja laiendatud, need on asjakohased; näidetest kasvavad loogiliselt välja lõikude järeldused; tekst on arutlev.\\
2 if: alaväited selgelt sõnastatud ja osaliselt seotud põhiväitega; alaväited esitatud loogilises järjekorras; kõiki näiteid selgitatud ja laiendatud, need on asjakohased; näidetest ei kasva selgelt välja järeldused; tekst on arutlev.\\
1 if: kõik alaväited pole selgelt sõnastatud; alaväited on osaliselt seotud põhiväitega; toodud näiteid osaliselt laiendatud; näidetest ei kasva välja järeldused; tekstil on valdavalt jutustav/kirjeldav iseloom; esineb üksikud kergemad faktivead.\\
0 if: argumentatsioon on seosetu või puudub, tekst on jutustav/kirjeldav; esineb küsitavusi/faktivead.\\
\{Output instructions\}

\{Preface\} Lausemoodustus (ühildumine, sõnajärg, rektsioon). \{Grading instructions\}\\
3 if: laused on arusaadavad ja terviklikud; kasutatakse sidusaid ja erineva ülesehitusega lauseid; lausestuseksimusi ei esine.\\
2 if: laused arusaadavad ja terviklikud; kasutatakse sidusaid ja sarnase ülesehitusega lauseid; esineb üksikuid lausestuseksimusi.\\
1 if: laused suuremalt jaolt arusaadavad; kasutatakse suuremalt jaolt sidusaid lauseid, mille ülesehitus ühekülgne; esineb palju lausestuseksimusi.\\
0 if: laused ebaselged ja välja arendamata, lausestuseksimuste tõttu tekst arusaamatu.\\
\{Output instructions\}

\{Preface\} Lõpetus. \{Grading instructions\}\\
3 if: sissejuhatuses püstitatud probleemile on vastatud; lõikude peamised järeldused on teises sõnastuses kokku võetud.\\
2 if: sissejuhatuses püstitatud probleemile on vastatud; lõikude peamisi järeldusi korratakse sissejuhatusele ja teemaarendusele ligilähedases sõnastuses.\\
1 if: sissejuhatuses püstitatud probleemile on vastatud osaliselt või tuuakse sisse uus teema; lõikude peamisi järeldusi korratakse sissejuhatuse ja teemaarendusega samas sõnastuses.\\
0 if: lõpetus pole teemaarenduse või sissejuhatusega seotud.\\
\{Output instructions\}

\{Preface\} Kirjavahemärgistus. \{Grading instructions\}\\
3 if: kirjavahemärgistus on täpne; võib esineda 0-1 viga kokku.\\
2 if: kirjavahemärgistus valdavalt täpne; võib esineda 2-3 viga kokku.\\
1 if: kirjavahemärgistuses esineb palju puudusi; võib esineda 4-5 viga kokku.\\
0 if: kirjavahemärgistus puudulik.\\
This aspect only refers to the correct usage of any punctuation like koma, jutumärgid, kriipsud, koolon, lauselõpumärgid, if and where relevant (but NOT lack of space after punctuation). Do not count repeated mistakes of the same type as new mistakes (only reduce 1 point per similar mistake type).\\
\{Output instructions\}

\{Preface\} Õigekiri ja vormimoodustus. \{Grading instructions\}\\
3 if: õigekiri ja vormimoodustus on korrektne; võib esineda 0-1 viga kokku.\\
2 if: valdavalt korrektne; 2-3 viga kokku.\\
1 if: palju puudusi; 4-5 viga kokku.\\
0 if: õigekiri ja vormimoodustus puudulik, 6 või enam viga.\\
This aspect only considers algustähed, sõnade kokku- ja lahkukirjutamine, häälikuortograafia, käänamine-pööramine. Ignore obvious typos like accidentally swapped letters in a word. Do not count repeated mistakes of the same type as new mistakes, only reduce 1 point per mistake type. The following count as one mistake even if repeated: 1) ühes sõnas esinevad eri tüüpi õigekeelsusvead; 2) samas sõnas või morfoloogilises vormis korduvad vead; 3) sama lausetarindi puhul korduvalt ära jäetud või ülearune kirjavahemärk või kahepoolne vahemärk (jutumärgid, sulud); 4) sama/sarnase kõrvallause, lauselühendi, kiilu, sh ütte ja hüüundi, järellisandi ja -täiendi ning lisanditaolise määruse puhul korduvalt ära jäetud koma(d) või mõttekriips(ud), Näiteks loetakse üheks veaks 1) eksimused veaohtlikus sõnatüübis sama vormi moodustamisel, nt astmevahelduslike ikliiteliste sõnade sama käändevormi moodustamise vead (nt osastava vormid *mõistliku, *tuleviku); vead kontsert-tüüpi võõrsõnade käänamisel ning pesa- ja kõne-tüüpi sõnade osastavavormide moodustamisel (nt *kontsertile, *asfaltile; austas *pere, kuulas *kõne); 2) eksimused likkus-liitelistes sõnades käändevormist sõltumata (nt omastav *avalikuse, nimetav *tegelikus); 3) si-vorm mitmuse osastavas (nt *tublisi, *külasi, *vendasi) muuttüübist sõltumata; 4) sama lausetüübi puhul korduvalt ära jäetud lõpumärk; 5) samas tarindis või lauseümbruses koma(de) puudumine, nt põimlauses sarnases ümbruses (kuhu-/kus-/kust-kohakõrvallaused; täiendkõrvallaused sõltumata siduvast asesõnast ja selle käändevormist (nt mis-, mille-, mida-, millega-; kes-, kellele-, kellest-vormiga algavad täiendkõrvallaused))\\
\{Output instructions\}

}

\subsection*{A3. Linguistic features used for supervised learning}

In our supervised learning experiments, we applied various types of linguistic features previously used in automatic assessment of Estonian L2 learner writings. \textcite{vajjala-loo-2014-automatic} relied on lexical and morphological features, while \textcite{allkivi-2025-cefr} also included surface complexity and error features to classify texts on the scale of A2--C1.

For tokenization, sentence segmentation, lemmatization, and morphological tagging of essays, we used the Stanza Python package \parencite{qi-etal-2020-stanza}. Error frequencies were calculated based on the output of a Llama 2 model fine-tuned for Estonian grammatical error correction \parencite{luhtaru_err_2024} and a context-sensitive statistical spell-checker \parencite{allkivi-metsoja-kippar-2023-spelling}. Further feature extraction and data processing were done with the Pandas package for Python \parencite{mckinney2010}. We could largely reuse the feature extraction scripts developed for the assessment of Estonian L2 texts\footnote{\url{https://github.com/tlu-dt-nlp/Estonian-CEFR-Assessment/}} (see Allkivi et al. \citeyear{allkivi_elle_2024}; Allkivi \citeyear{allkivi-2025-cefr}). However, we added new features that characterize the text structure and made use of a more advanced grammar correction tool together with error type classification.

\textbf{Error features} involve 10 types of automated edits detected by comparing the original and corrected text. More generally, the edits can be categorized as word-level, punctuation, whitespace, i.e., compounding, and word order corrections. Words and punctuation marks can be replaced, added, or deleted. In addition, the category of mixed corrections is formed by cases where several edit types co-occur, e.g., a word is replaced or a punctuation mark added within the scope of a word order correction. Word replacements occur due to spelling, word form, or lexical choice errors. Such corrections could not be automatically distinguished at the time of the analysis, but we used the separate spell-checking tool to estimate the frequency of spelling errors.

For each correction category, we determined the number of respective edits as well as their ratio to all edits made by the grammar corrector, or the ratio of corrected words to the total number of words in the case of spelling corrections. Such features convey the distribution of various types of corrections in text. Furthermore, we computed three general features based on grammar correction output: the total count of corrections, and the number of corrections per word and sentence.

\textbf{Surface features} include text complexity metrics and text structure variables that do not require deeper morphosyntactic analysis but can be calculated based on word, sentence, and paragraph segmentation and syllabification. Features, such as average word length in characters, average sentence length in words, LIX index \parencite{bjornsson-lix}, SMOG index \parencite{mclaughlin-smog}, and Flesch-Kincaid Grade Level index \parencite{kincaid-etal}, characterize readability, i.e., how easy it is to understand written text. This is stated as one of the criteria for grading \emph{syntax}. Word length can also be associated with lexical complexity and thus the aspect of \emph{vocabulary}.

To score \emph{structuring and formatting}, we used measures of paragraph count and length. According to the scoring rubric, the text should be divided into proportional paragraphs. We calculated the mean, standard deviation, and maximum difference in word and sentence count within paragraphs.

\textbf{Lexical features} represent three dimensions of lexical complexity: diversity, sophistication, and density. Lexical diversity is most simply captured by the number of unique words (lemmas), and their ratio to the total word count, referred to as the type-token ratio (TTR). We also used various standardized indices that reduce the impact of text length: root type-token ratio (RTTR; Guiraud \citeyear{guiraud-rttr}), Maas index \parencite{maas-diversity}, Uber index \parencite{dugast-uber}, and the Measure of Textual Lexical Diversity (MTLD) proposed by \textcite{mccarthy-jarvis-2010}. Following \textcite{lu-2012-lexical}, \textcite{vajjala-loo-2014-automatic}, and \textcite{allkivi-2025-cefr}, we added part-of-speech-specific TTRs and corrected verb variation (CVV) to the feature set.

Lexical sophistication is mostly defined by the rarity, but also the abstractness/concreteness of vocabulary \parencite{brysbaert-etal-abstractness}. We compared the student essays to the lemmatized version of the Estonian frequency dictionary\footnote{\url{https://www.cl.ut.ee/ressursid/sagedused1/index.php?lang=en}}, using separate tiers of top-frequency words (1,000, 2,000, 3,000, 4,000, and 5,000 most frequent lemmas) to count the proportion of advanced tokens. In addition, we rated average noun abstractness with an Estonian speed-reading software tool\footnote{\url{https://kiirlugemine.keeleressursid.ee}}. The rating of noun tokens is based on the annotated dataset of \textcite{mikk-etal-2003} and a three-point scale, where abstract nouns that designate objects and notions imperceptible by senses are graded as 3 (see Mikk \& Elts \citeyear{mikk-elts-1993}).

\textbf{Grammatical features} are divided into part of speech (POS), nominal, and verb features. POS features comprise the ratio of nouns, pronouns, adjectives, verbs, adverbs, conjunctions, adpositions, and interjections. We also extracted the ratio of some POS subcategories, such as the different types of pronouns (personal, reflexive, interrogative-relative, demonstrative, and indefinite), conjunctions (subordinating vs. coordinating), and adpositions (postpositions vs. prepositions). Two other POS-based features – noun-to-verb ratio and formality index (F-score) introduced by \textcite{heylighen-dewaele-2002} – reflect readability and have proven to distinguish text genres and written texts from oral speech in Estonian \parencite{kerge-pajupuu-2010, pajupuu-etal-2010, puksand-kerge-2011}.

Nominal features include the number of noun, pronoun, and adjective case forms represented in the text; the ratio of plural nominals as well as plural forms of nouns, pronouns, and adjectives; the ratio of the different case forms: nominative, genitive, partitive, additive, illative, inessive, elative, allative, adessive, ablative, translative, terminative, essive, abessive, and comitative case. Unlike \textcite{allkivi-2025-cefr}, we only considered the total frequency of case forms and ignored the POS differences, as we did not expect the distribution of nominal cases to impact the scores of L1 essays as greatly as it affects the grading of L2 essays.

Verb features contain the ratio of indicative, conditional, and imperative mood forms; finite and nonfinite forms, specifically infinitives, participles, and gerunds; present and past tense forms; singular and plural forms; forms of passive voice; and words that constitute negative verb forms.

We associated each feature with one or more aspects in the scoring rubric and used the selected features to train prediction models. While all aspects relate to linguistic accuracy, grading \emph{punctuation} and \emph{orthography and morphology} relied on error features exclusively. Scoring \emph{structuring and formatting} also required surface features characterizing text structure. \emph{Syntax} was rated using grammatical features, as well as surface complexity, and related error and lexical features. The score of \emph{vocabulary} was predicted based on lexical features, POS features, some error features, and word length. Features overlapping between rubric categories were either ambiguous correction types (e.g., spelling, word replacement, and mixed errors) or POS features, which are lexicogrammatical in nature.

All in all, 6 features were connected to \emph{punctuation}, 11 features to \emph{orthography and morphology} as well as \emph{structuring and formatting}, 52 features to \emph{vocabulary}, and 62 features to \emph{syntax}. Table \ref{table_corr} presents the features most strongly correlated with each scoring aspect, highlighting the differences between the 9th and 12th grades. However, more features were included in the automatic assessment. For example, \emph{syntax} was also scored based on mixed correction ratio, and a number of other features in the case of 12th grade. These contained error features, such as missing and unnecessary word corrections, as well as verb features (2nd and 1st person verb forms, indicative, imperative, and the overall ratio of finite verb forms) and nominal features (adjective and noun ratio, the number of pronoun case forms). Lexical density was also relevant for grading \emph{vocabulary}, along with TTR in 9th grade, and replaced word corrections, pronoun, and adjective ratios in 12th grade.

\begin{table}[ht]
\caption{Best predictor features based on the correlation with respective subscore. Pearson correlation coefficients have been averaged across the training sets of 10 cross-validation folds. The standard deviations are given in brackets. The table includes up to 10 most correlated features per category or all features involved in the optimal regression pipeline. Features that were removed from supervised learning experiments due to multicollinearity are marked with an asterisk. Features that were only significant in grading 9th or 12th grade essays are marked in bold.}
\label{table_corr}
\begin{minipage}[t]{0.48\textwidth}
\vspace{0pt}
\centering
\begin{tabular}{llrlr}
Category & 9th grade & & 12th grade & \\
  \hline
  Punctuation & (1) Missing punct. corrections & -.75 (.01) & (1) Missing punct. corrections & -.59 (.01) \\
  & (2) Missing punct. correction ratio* & -.66 (.01) & (2) Missing punct. correction ratio* & -.56 (.01) \\
  & (3) Unnecessary punct. corrections & -.13 (.01) & (3) Unnecessary punct. corrections & -.24 (.01) \\
  & (4) Replaced punct. corrections & -.13 (.01) & (4) Replaced punct. corrections & -.10 (.01) \\
  \hline
  Orthography & (1) Total errors per word & -.69 (.01) & (1) Total errors per word & -.64 (.01) \\
  \& morphology & (2) Total error count* & -.67 (.01) & (2) Total error count* & -.57 (.01) \\
  & (3) \textbf{Spell-corrected word ratio} & -.57 (.01) & (3) Total errors per sentence* & -.51 (.01) \\
  & (4) \textbf{Spelling corrections*} & -.57 (.01) & (4) Replaced word corrections & -.46 (.01) \\
  & (5) Replaced word corrections & -.57 (.01) & (5) Mixed corrections & -.42 (.01) \\
  & (6) Total errors per sentence* & -.57 (.01) & (6) Whitespace corrections & -.38 (.01) \\
  & (7) Mixed corrections & -.56 (.01) & & \\
  & (8) \textbf{Mixed correction ratio} & -.31 (.01) & & \\
  & (9) Whitespace corrections & -.31 (.01)  & & \\
  \hline
  Structuring & (1) Spell-corrected word ratio & -,37 (.01) & (1) Spell-corrected word ratio & -.26 (.02) \\
  \& formatting & (2) Spelling corrections* & -.36 (.01) & (2) Replaced word corrections & -.24 (.01) \\
  & (3) SD of paragraph word count & -.35 (.01) & (3) Diff. of paragraph word count & -.19 (.01) \\
  & (4) Diff. of paragraph word count* & -.34 (.01) & (4) SD of paragraph word count* & -.19 (.01) \\
  & (5) Replaced word corrections & -.29 (.01) & (5) Spelling errors* & -.18 (.01) \\
  & (6) SD of paragraph sentence count & -.22 (.01) & (6) Mean paragraph sentence count & .15 (.01) \\
  & (7) Diff. of paragraph sentence count* & -.18 (.01) & (7) SD of paragraph sentence count & -.13 (.01) \\
  & (8) \textbf{Replaced word correction ratio} & .14 (.01) & (8) \textbf{Paragraph count} & -.12 (.02) \\
  & (9) Mean paragraph word count & -.12 (.01) & (9) Diff. of paragraph sentence count* & -.12 (.01) \\
  & (10) Mean paragraph sentence count & .11 (.02) & (10) Mean paragraph word count & .11 (.02) \\
  \hline
  Vocabulary & (1) Mixed corrections & -.42 (.01) & (1) \textbf{Lemma count} & .43 (.01) \\
  & (2) Replaced word corrections & -.39 (.01) & (2) RTTR* & .40 (.01) \\
  & (3) Mean word length & .36 (.02) & (3) \textbf{CVV} & .32 (.01) \\
  & (4) RTTR & .32 (.01) & (4) MTLD & .30 (.01) \\
  & (5) Maas index* & -.30 (.01) & (5) Mixed corrections & -.29 (.01) \\
  & (6) MTLD & .30 (.01) & (6) Mean word length & .24 (.01) \\
  & (7) \textbf{Rare word ratio (beyond 1,000)} & .30 (.02) & (7) Personal pronoun ratio & -.23 (.01) \\
  & (8) Uber index* & .29 (.01) & (8) \textbf{2nd person verb form ratio} & -.22 (.02) \\
  & (9) \textbf{Adjective ratio} & .28 (.01) & (9) Maas index & -.22 (.01) \\
  & (10) \textbf{Rare word ratio (beyond 5,000)} & .26 (.02) & (10) \textbf{Rare word ratio (beyond 3,000)} & .21 (.01) \\
  \hline
  Syntax & (1) Mixed corrections & -.46 (.01) & (1) Mixed corrections & -.34 (.01) \\
  & (2) Mean word length & .40 (.02) & (2) \textbf{No. of noun case forms} & .26 (.01) \\
  & (3) \textbf{Rare word ratio (beyond 1,000)} & .34 (.02) & (3) \textbf{Total no. of nominal case forms*} & .26 (.01) \\
  & (4) Adjective ratio & .32 (.01) & (4) Mean word length & .24 (.02) \\
  & (5) \textbf{Rare word ratio (beyond 5,000)} & .30 (.02) & (5) Lexical density & .23 (.01) \\
  & (6) \textbf{Mean sentence length} & -.29 (.01) & (6) No. of adjective case forms & .23 (.01) \\
  & (7) \textbf{Flesch-Kincaid Grade Level*} & .29 (.01) & (7) \textbf{Rare word ratio (beyond 3,000)} & .22 (.01) \\
  & (8) No. of adjective case forms & .26 (.01) & (8) \textbf{Pronoun ratio} & -.21 (.01) \\
  & (9) Lexical density & .26 (.02) & (9) \textbf{LIX index} & .21 (.01) \\
  & (10) \textbf{F-index*} & .25 (.01) & (10) \textbf{Word order errors} & -.20 (.01) \\
   \hline
\end{tabular}
\end{minipage}
\end{table}

\end{document}